%% file: main.tex
\pdfoutput=1
\documentclass[12pt,a4paper,oneside]{book}

\usepackage{amsmath,amsthm,amssymb,graphicx,hyperref}
\usepackage[left=1.2in,right=1.2in,top=1in,bottom=1in]{geometry}
\usepackage[english]{babel}

\usepackage{parskip}

\usepackage{xcolor}
\usepackage[frozencache,cachedir=minted-cache]{minted}
\usepackage{dirtree}

\usepackage{paralist}
\RequirePackage{hyphenat}

\usepackage[ruled]{algorithm2e}
\SetKwComment{Comment}{$\triangleright$\ }{}

\usepackage{rotating}
\usepackage{multicol}
\usepackage{subfig}
\usepackage{caption}
\usepackage[acronym,toc,nomain]{glossaries}
\usepackage{glossary-mcols}

\loadglsentries{sections/glossary}
\makeglossaries
\usepackage{appendix}
\usepackage[inline]{enumitem}

\graphicspath{{./images/}}

\begin{document}
\sloppy

\thispagestyle{empty}

\begin{center}
  \begin{figure}[h!]
    \vspace{-20pt}
    \begin{center}
      \includegraphics[width=100pt]{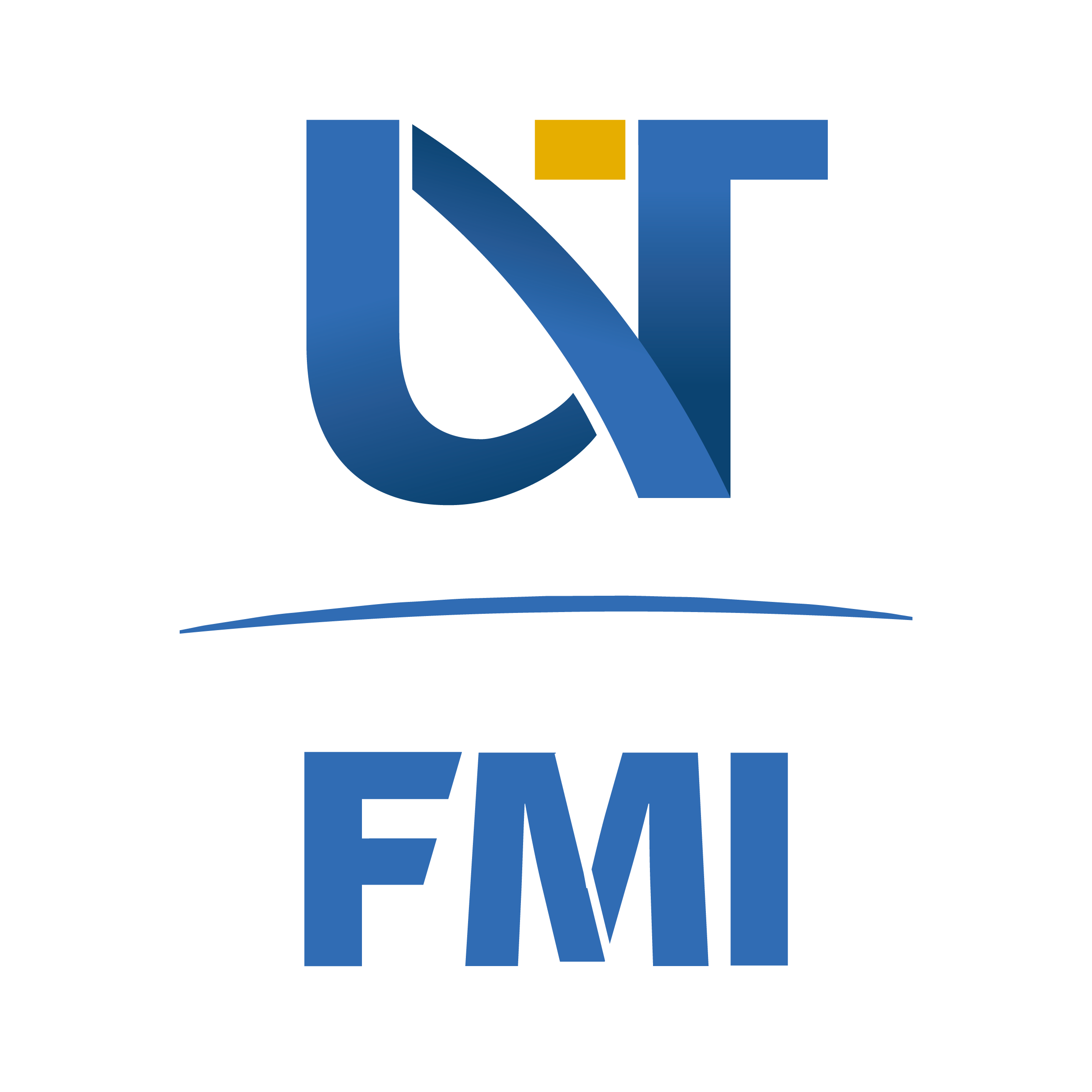}
    \end{center}
\end{figure}

{\large{\bf WEST UNIVERSITY OF TIMIŞOARA\\FACULTY OF MATHEMATICS AND COMPUTER SCIENCE\\MASTER STUDY PROGRAM:\\Artificial Intelligence and Distributed Computing}}

\vspace{120pt}
{\huge {\bf MASTER THESIS}}

\vspace{160pt}
\end{center}

{\large\noindent{\bf SUPERVISOR:\hfill GRADUATE:}\\ \noindent Lect. Dr. Marian Neagul\hfill \noindent Alexandru-Constantin Munteanu}
\vfill

\begin{center}
  {\bf TIMIŞOARA\\2021}
\end{center}

\newpage
\thispagestyle{empty}
\begin{center}
  {\large{\bf WEST UNIVERSITY OF TIMIŞOARA\\FACULTY OF MATHEMATICS AND COMPUTER SCIENCE\\MASTER STUDY PROGRAM:\\Artificial Intelligence and Distributed Computing}}
  \vspace{200pt}
  
  {\huge {\bf Semantic Segmentation of Vegetation in Remote Sensing Imagery Using Deep Learning }}
  \vspace{180pt}
\end{center}

{\large\noindent{\bf SUPERVISOR:\hfill GRADUATE:} \\ \noindent Lect. Dr. Marian Neagul\hfill \noindent Alexandru-Constantin Munteanu}

\vfill
\begin{center}
  {\bf TIMIŞOARA\\2021}
\end{center}

\newpage
\normalsize{}

\input{sections/abstract}


\tableofcontents

\input{sections/introduction}

\input{sections/sota}
\input{sections/approach}
\input{sections/experiments}
\input{sections/conclusions}

\bibliography{main}
\bibliographystyle{acm}
\addcontentsline{toc}{chapter}{Bibliography}
\printglossary[style=mcolindex, title=List of Abbreviations, nonumberlist]
\input{sections/appendix}

\end{document}

%% file: sections/glossary.tex

\newglossaryentry{adt}{name=ADT, description={Automatic Drum Transcription}}
\newglossaryentry{ae}{name=AE, description={AutoEncoder}}
\newglossaryentry{agi}{name=AGI, description={Artificial General Intelligence}}
\newglossaryentry{ai}{name=AI, description={Artificial Intelligence}}
\newglossaryentry{amt}{name=AMT, description={Automatic Music Transcription}}
\newglossaryentry{ann}{name=ANN, description={Artificial Neural Network}}
\newglossaryentry{arnn}{name=ARNN, description={Anticipation Recurrent Neural Network}}
\newglossaryentry{bilstm}{name=BILSTM, description={Bidirectional Long Short-Term Memory}}
\newglossaryentry{bptt}{name=BPTT, description={Back-Propagation Through Time}}
\newglossaryentry{brnn}{name=BRNN, description={Bidirectional Recurrent Neural Network}}
\newglossaryentry{bqml}{name=BQML, description={Big Query Machine Learning}}
\newglossaryentry{cdbn}{name=CDBN, description={Convolutional Deep Belief Networks}}
\newglossaryentry{cec}{name=CEC, description={Constant Error Carousel}}
\newglossaryentry{clnn}{name=CLNN, description={ConditionaL Neural Networks}}
\newglossaryentry{cnn}{name=CNN, description={Convolutional Neural Network}}
\newglossaryentry{cv}{name=CV, description={Computer Vision}}
\newglossaryentry{convnet}{name=ConvNet, description={Convolutional Neural Network}}
\newglossaryentry{crbm}{name=CRBM, description={Conditional Restricted Boltzmann Machine}}
\newglossaryentry{crnn}{name=CRNN, description={Convolutional Recurrent Neural Network}}
\newglossaryentry{dae}{name=DAE, description={Denoising AutoEncoder or Deep AutoEncoder}}
\newglossaryentry{dbm}{name=DBM, description={Deep Boltzmann Machine}}
\newglossaryentry{dbn}{name=DBN, description={Deep Belief Network}}
\newglossaryentry{dl}{name=DL, description={Deep Learning}}
\newglossaryentry{dnn}{name=DNN, description={Deep Neural Network}}
\newglossaryentry{dsn}{name=DSN, description={Deep Stacking Network}}
\newglossaryentry{dwt}{name=DWT, description={Discrete Wavelet Transform}}
\newglossaryentry{elm}{name=ELM, description={Extreme Learning Machine}}
\newglossaryentry{etl}{name=ETL, description={Extract, Transform, Load Pipeline}}
\newglossaryentry{fc}{name=FC, description={Fully Connected}}
\newglossaryentry{fcn}{name=FCN, description={Fully Convolutional Network}}
\newglossaryentry{fc-cnn}{name=FC-CNN, description={Fully Convolutional Convolutional Neural Network}}
\newglossaryentry{fc-lstm}{name=FC-LSTM, description={Fully Connected Long Short-Term Memory}}
\newglossaryentry{gan}{name=GAN, description={Generative Adversarial Network}}
\newglossaryentry{gbrcn}{name=GBRCN, description={Gradient-Boosting Random Convolutional Network}}
\newglossaryentry{gfnn}{name=GFNN, description={Gradient Frequency Neural Networks}}
\newglossaryentry{glcm}{name=GLCM, description={Gray Level Co-occurrence Matrix}}
\newglossaryentry{han}{name=HAN, description={Hierarchical Attention Network}}
\newglossaryentry{hhds}{name=HHDS, description={HipHop Dataset}}
\newglossaryentry{lstm}{name=LSTM, description={Long Short-Term Memory}}
\newglossaryentry{mclnn}{name=MCLNN, description={Masked ConditionaL Neural Networks}}
\newglossaryentry{mer}{name=MER, description={Music Emotion Recognition}}
\newglossaryentry{ml}{name=ML, description={Machine Learning}}
\newglossaryentry{mlm}{name=MLM, description={Music Language Models}}
\newglossaryentry{mlp}{name=MLP, description={Multi-Layer Perceptron}}
\newglossaryentry{mrs}{name=MRS, description={Music Recommender System}}
\newglossaryentry{msdae}{name=MSDAE, description={Modified Sparse Denoising Autoencoder}}
\newglossaryentry{mse}{name=MSE, description={Mean Squared Error}}
\newglossaryentry{msr}{name=MSR, description={Music Style Recognition}}
\newglossaryentry{nn}{name=NN, description={Neural Network}}
\newglossaryentry{nnmodff}{name=NNMODFF, description={Neural Network based Multi-Onset Detection Function Fusion}}
\newglossaryentry{odf}{name=ODF, description={Onset Detection Function}}
\newglossaryentry{pnn}{name=PNN, description={Probabilistic Neural Network}}
\newglossaryentry{poc}{name=POC, description={Proof of Concept}}
\newglossaryentry{prelu}{name=PReLU, description={Parametric Rectified Linear Unit}}
\newglossaryentry{ransac}{name=RANSAC, description={RANdom SAmple Consensus}}
\newglossaryentry{rbm}{name=RBM, description={Restricted Boltzmann Machine}}
\newglossaryentry{relu}{name=ReLU, description={Rectified Linear Unit}}
\newglossaryentry{ricnn}{name=RICNN, description={Rotation Invariant Convolutional Neural Network}}
\newglossaryentry{rnn}{name=RNN, description={Recurrent Neural Network}}
\newglossaryentry{rtrl}{name=RTRL, description={Real-Time Recurrent Learning}}
\newglossaryentry{sae}{name=SAE, description={Stacked AE}}
\newglossaryentry{sdae}{name=SDAE, description={Stacked DAE}}
\newglossaryentry{sgd}{name=SGD, description={Stochastic Gradient Descent}}
\newglossaryentry{svm}{name=SVM, description={Support Vector Machine}}
\newglossaryentry{svd}{name=SVD, description={Singing Voice Detection}}
\newglossaryentry{svs}{name=SVS, description={Singing Voice Separation}}
\newglossaryentry{vad}{name=VAD, description={Voice Activity Detection}}
\newglossaryentry{vae}{name=VAE, description={Variational AutoEncoder}}
\newglossaryentry{vpnn}{name=VPNN, description={Vector Product Neural Network}}
\newglossaryentry{wpe}{name=WPE, description={Weighted Prediction Error}}


\newglossaryentry{gis}{name=GIS, description={Geographic Information Systems}}
\newglossaryentry{aecoo}{name=AECOO, description={Architecture, Engineering, Construction, Owner and Operator}}
\newglossaryentry{agile}{name=AGILE, description={Association of Geographic Information Laboratories in Europe}}
\newglossaryentry{ansi}{name=ANSI, description={American National Standards Institute}}
\newglossaryentry{arml}{name=ARML, description={Augmented Reality Markup Language}}
\newglossaryentry{as}{name=AS, description={Abstract Specification}}
\newglossaryentry{ascii}{name=ASCII, description={American Standard Code for Information Interchange}}
\newglossaryentry{bmp}{name=BMP, description={Bitmap. A Microsoft Windows image format}}
\newglossaryentry{bp}{name=BP, description={Best Practice}}
\newglossaryentry{cad}{name=CAD, description={Computer-aided design and drafting}}
\newglossaryentry{cen}{name=CEN, description={Comite Europeen de Normalisation (European Committee for Standardization)}}
\newglossaryentry{cgdi}{name=CGDI, description={Canadian Geospatial Data Infrastructure}}
\newglossaryentry{cipi}{name=CIPI, description={A 2002 OGC Interoperability Initiative focused on critical infrastructure protection}}
\newglossaryentry{cite}{name=CITE, description={OGC Compliance and Interoperability Testing Initiative, the OGC's Compliance Testing Program.}}
\newglossaryentry{citygml}{name=CityGML, description={OGC City Geography Markup Language (CityGML) Encoding Standard}}
\newglossaryentry{com}{name=COM, description={Common Object Model, a Microsoft paradigm to connect components}}
\newglossaryentry{corba}{name=CORBA, description={Common Object Request Broker Architecture}}
\newglossaryentry{crs}{name=CRS, description={Coordinate Reference System. Also CRS Domain Working Group.}}
\newglossaryentry{cs}{name=CS, description={OGC Catalogue Service Interface Standard}}
\newglossaryentry{csw}{name=CSW, description={OGC Catalog Service Interface Standard for the Web}}
\newglossaryentry{ct}{name=CT, description={OGC Coordinate Transformation Encoding Standard}}
\newglossaryentry{ctl}{name=CTL, description={Compliance Test Language}}
\newglossaryentry{cuahsi}{name=CUAHSI, description={Consortium of Universities for the Advancement of Hydrologic Science}}
\newglossaryentry{dbf}{name=DBF, description={Data Base File, the Dbase file format}}
\newglossaryentry{dbms}{name=DBMS, description={Data Base Management System}}
\newglossaryentry{dce}{name=DCE, description={Distributed Computing Environment}}
\newglossaryentry{dcp}{name=DCP, description={Distributed Computing Platform}}
\newglossaryentry{dem}{name=DEM, description={Digital Elevation Model}}
\newglossaryentry{demts}{name=DEMTS, description={Digital and Electronic Maps Transfer Standard (Russia)}}
\newglossaryentry{dgiwg}{name=DGIWG, description={Digital Geospatial Information Working Group}}
\newglossaryentry{dgn}{name=DGN, description={DesiGN file, the Microstation drawing format}}
\newglossaryentry{digest}{name=DIGEST, description={Digital Geographic Exchange Standard}}
\newglossaryentry{dlg}{name=DLG, description={Digital Line Graph}}
\newglossaryentry{dlm}{name=DLM, description={Digital Landscape Model}}
\newglossaryentry{dp}{name=DP, description={Discussion Paper}}
\newglossaryentry{dtm}{name=DTM, description={Digital Terrain Model}}
\newglossaryentry{dwg}{name=DWG, description={Domain Working Group}}
\newglossaryentry{dxf}{name=DXF, description={Drawing eXchange Format (AutoCAD exchange format)}}
\newglossaryentry{ea}{name=EA, description={Enterprise Architecture}}
\newglossaryentry{easig}{name=EASIG, description={Enterprise Architecture Special Interest Group}}
\newglossaryentry{ebrim}{name=ebRIM, description={OASIS ebXML (OASIS Electronic Business using eXtensible Markup Language) Registry Information Model. See OGC Catalog ebRIM Application Profile: Earth Observation Products.}}
\newglossaryentry{eea}{name=EEA, description={European Environment Agency}}
\newglossaryentry{eo}{name=EO, description={Earth Observation}}
\newglossaryentry{eo2heaven}{name=EO2heaven, description={Earth Observation and Environmental Modelling for the Mitigation of Health Risks}}
\newglossaryentry{eosdis}{name=EOSDIS, description={Earth Observing System Data and Information System}}
\newglossaryentry{epsg}{name=EPSG, description={European Petroleum Survey GroupEuroSDR European Spatial Data Research Network}}
\newglossaryentry{esdi}{name=ESDI, description={European Spatial Data Information Infrastructure}}
\newglossaryentry{eurogi}{name=EUROGI, description={European Umbrella Organisation for Geographic Information}}
\newglossaryentry{geoapi}{name=GeoAPI, description={OGC GeoAPI Interface Standard (a Java application programming interface)}}
\newglossaryentry{geoxacml}{name=GeoXACML, description={OGC Geospatial eXtensible Access Control Markup Language Standard}}
\newglossaryentry{getis}{name=GETIS, description={Geoprocessing networks in a European Territorial Interoperability Study}}
\newglossaryentry{gita}{name=GITA, description={Geospatial Information \& Technology Association}}
\newglossaryentry{gml}{name=GML, description={OGC Geography Markup Language Encoding Standard}}
\newglossaryentry{go}{name=GO, description={Geospatial Objects (a retired OGC standard)}}
\newglossaryentry{gpc/gis}{name=GPC/GIS, description={GPC Global Information Solutions}}
\newglossaryentry{gps}{name=GPS, description={Global Positioning System}}
\newglossaryentry{grss}{name=GRSS, description={IEEE Geoscience and Remote Sensing Society}}
\newglossaryentry{gsdi}{name=GSDI, description={Global Spatial Data Infrastructure}}
\newglossaryentry{gservrestswg}{name=GServRestSWG, description={GeoServices REST (REpresentational State Transfer) Standards Working Group}}
\newglossaryentry{hdf}{name=HDF, description={Hierarchical Data Format}}
\newglossaryentry{hti}{name=HTI, description={Human Technology Interface}}
\newglossaryentry{ics}{name=ICS, description={Information Communities and Semantics}}
\newglossaryentry{ict}{name=ICT, description={Information and Communications Technology}}
\newglossaryentry{ie}{name=IE, description={Interoperability Experiment}}
\newglossaryentry{iec}{name=IEC, description={ISO IEC JTC 1/SC 24/WG 8 Computer graphics, image processing and environmental presentation}}
\newglossaryentry{iemss}{name=iEMSs, description={International Modeling \& Software Society Secretariat}}
\newglossaryentry{ies}{name=IES, description={Imagery Exploitation Working Group IETF, description={Internet Engineering Task Force}}
\newglossaryentry{ietf}{name=IETF, description={Internet Engineering Task Force}}
\newglossaryentry{ilaf}{name=ILAF, description={OGC Iberian and Latin-American Forum}}
\newglossaryentry{indoorgml}{name=IndoorGML, description={common schema framework for interoperability between indoor and outdoor navigation applications, being drafted by the OGC IndoorGML Standards Working Group}}
\newglossaryentry{ip}{name=IP, description={Depending on context, either Intellectual Property or Interoperability Program}}
\newglossaryentry{ipr}{name=IPR Intellectual Property Rights}}
\newglossaryentry{ipteam}{name=IPTeam, description={Interoperability Program Team}}
\newglossaryentry{is}{name=IS, description={Implementation Standard}}
\newglossaryentry{iso}{name=ISO, description={International Organization for Standardization}}
\newglossaryentry{isoc}{name=ISOC, description={Internet Society}}
\newglossaryentry{isprs}{name=ISPRS, description={International Society for Photogrammetry and Remote Sensing}}
\newglossaryentry{itu}{name=ITU, description={International Telecommunication Union}}
\newglossaryentry{jag}{name=JAG, description={Joint Advisory Committee between OGC and ISO TC 211}}
\newglossaryentry{jpeg}{name=JPEG, description={Joint Photographics Group}}
\newglossaryentry{kml}{name=KML, description={OGC KML Encoding Standard (formerly Keyhole Markup Language)}}
\newglossaryentry{lbs}{name=LBS, description={Location Based Services}}
\newglossaryentry{los}{name=LOS, description={Location Organizer Folder}}
\newglossaryentry{mcp}{name=MCP, description={OGC Marketing \& Communications Program}}
\newglossaryentry{mismo}{name=MISMO, description={Mortgage Information Standards Maintenance Organization}}
\newglossaryentry{mmi}{name=MMI, description={OGC`s Multi-Hazard Mapping Initiative (MMI) Phase I (2001)}}
\newglossaryentry{mpp}{name=MPP, description={OGC`s Military Pilot Project (MPP) (2001)}}
\newglossaryentry{ncoic}{name=NCOIC, description={Net Centric Operations Infrastructure Committee}}
\newglossaryentry{netcdf}{name=NetCDF, description={OGC Network Common Data Form (NetCDF)}}
\newglossaryentry{ngo}{name=NGO, description={Non-Governmental Organization}}
\newglossaryentry{nibs}{name=NIBS, description={National Institute for Building Standards}}
\newglossaryentry{nist}{name=NIST, description={National Institute of Standards and Technology}}
\newglossaryentry{nma}{name=NMA, description={National Mapping Agency}}
\newglossaryentry{nre}{name=NRE, description={Natural Resources and Environment}}
\newglossaryentry{nsdi}{name=NSDI, description={National Spatial Data Infrastructure}}
\newglossaryentry{ntf}{name=NTF, description={Neutral Transfer Format (in UK)}}
\newglossaryentry{oab}{name=OAB, description={OGC Architecture Board}}
\newglossaryentry{oasis}{name=OASIS, description={Organization for the Advancement of Structured Information Standards}}
\newglossaryentry{ocap}{name=OCAP, description={OGC Outreach and Community Adoption (now OGC Marketing and Communications)}}
\newglossaryentry{ogc}{name=OGC, description={Open Geospatial Consortium}}
\newglossaryentry{ogf}{name=OGF, description={Open Grid Forum}}
\newglossaryentry{ols}{name=OLS, description={OGC Location Services}}
\newglossaryentry{oma}{name=OMA, description={Open Mobile Alliance}}
\newglossaryentry{omg}{name=OMG, description={Object Management Group}}
\newglossaryentry{oo}{name=OO, description={Object Oriented}}
\newglossaryentry{geosms}{name=GeoSMS, description={OGC standard providing an extended Short Messaging Service (SMS) encoding and interface for communicating location content}}
\newglossaryentry{openls}{name=OpenLS, description={OGC Open Location Services Interface Standard}}
\newglossaryentry{osf}{name=OSF, description={Ordering Services Framework for Earth Observation Products}}
\newglossaryentry{orm}{name=ORM, description={OGC Reference Model}}
\newglossaryentry{oscre}{name=OSCRE, description={Open Standards Consortium for Real Estate}}
\newglossaryentry{osdm}{name=OSDM, description={Office of Spatial Data Management (Australia)}}
\newglossaryentry{osgeo}{name=OSGEO, description={Open Source Geospatial Foundation}}
\newglossaryentry{ows}{name=OWS, description={OGC Web Services}}
\newglossaryentry{owsc}{name=OWSC, description={OGC Web Services Common}}
\newglossaryentry{pc}{name=PC, description={Planning Committee}}
\newglossaryentry{png}{name=PNG, description={Portable Network Graphic}}
\newglossaryentry{puck}{name=PUCK, description={OGC standard defining a protocol for RS232 and Ethernet connected instruments}}
\newglossaryentry{rcm}{name=RCM, description={Risk and Crisis Management}}
\newglossaryentry{restful}{name=RESTful, description={REpresentational State Transfer programming style}}
\newglossaryentry{rfc}{name=RFC, description={Request for Comment}}
\newglossaryentry{rfi}{name=RFI, description={Request for Information}}
\newglossaryentry{rfp}{name=RFP, description={Request for Proposals (or Participation)}}
\newglossaryentry{rft}{name=RFT, description={Request for Technology}}
\newglossaryentry{rpc}{name=RPC, description={Remote Procedure Call}}
\newglossaryentry{rtd}{name=RTD, description={Research and Technology Development (Europe)}}
\newglossaryentry{sany}{name=SANY, description={Sensors Anywhere Consortium}}
\newglossaryentry{sar}{name=SAR, description={Synthetic Aperture Radar}}
\newglossaryentry{scots}{name=SCOTS, description={Standards based Commercial Off-The-Shelf software}}
\newglossaryentry{sdi}{name=SDI, description={Spatial Data Infrastructure}}
\newglossaryentry{sdts}{name=SDTS, description={Spatial Data Transfer Standard}}
\newglossaryentry{se}{name=SE, description={Symbology Encoding}}
\newglossaryentry{sensorml}{name=SensorML, description={OGC Sensor Model Language Encoding Standard}}
\newglossaryentry{sf}{name=SF, description={Simple Features}}
\newglossaryentry{sfs}{name=SFS, description={OGC Simple Features SQL Encoding Standard}}
\newglossaryentry{sif}{name=SIF, description={Standard Interchange Format}}
\newglossaryentry{sig}{name=SIG, description={Special Interest Group}}
\newglossaryentry{sld}{name=SLD, description={Styled Layer Descriptor}}
\newglossaryentry{smac}{name=SMAC, description={OGC Strategic Member Advisory Committee}}
\newglossaryentry{sme}{name=SME, description={Small or Medium Sized Enterprise}}
\newglossaryentry{soa}{name=SOA, description={Services Oriented Architecture}}
\newglossaryentry{soap}{name=SOAP, description={Simple Object Access Protocol}}
\newglossaryentry{sos}{name=SOS, description={OGC Sensor Observation Service Interface Standard}}
\newglossaryentry{sp}{name=SP, description={Standards Program}}
\newglossaryentry{sps}{name=SPS, description={OGC Sensor Planning Service Interface Standard}}
\newglossaryentry{sql}{name=SQL, description={Structured Query Language}}
\newglossaryentry{tiff}{name=TIFF, description={Tagged Image File Format}}
\newglossaryentry{svg}{name=SVG, description={Styled Vector Graphics}}
\newglossaryentry{swe}{name=SWE, description={OGC Sensor Web Enablement activity}}
\newglossaryentry{swg}{name=SWG, description={Standards Working Group}}
\newglossaryentry{tc}{name=TC, description={Technical Committee}}
\newglossaryentry{tdwg}{name=TDWG, description={Taxonomic Data Working Group}}
\newglossaryentry{tiger}{name=TIGER, description={Topologically Integrated Geographic Encoding and Reference file (US)}}
\newglossaryentry{tjs}{name=TJS, description={OGC Georeferenced Table Joining Service Encoding Standard}}
\newglossaryentry{tml}{name=TML, description={OGC Transducer Markup Language Encoding Standard (retired)}}
\newglossaryentry{ucar}{name=UCAR, description={University Corporation for Atmospheric Research / National Center for Atmospheric Research (NCAR)}}
\newglossaryentry{uddi}{name=UDDI, description={Universal Description, Discovery, and Integration}}
\newglossaryentry{uml}{name=UML, description={Unified Modeling Language}}
\newglossaryentry{usgif}{name=USGIF, description={US Geospatial Intelligence Foundation}}
\newglossaryentry{vr}{name=VR, description={Virtual Reality}}
\newglossaryentry{w3c}{name=W3C, description={World Wide Web Consortium}}
\newglossaryentry{wcps}{name=WCPS, description={OGC Web Coverage Processing Service Interface Standard}}
\newglossaryentry{wcs}{name=WCS, description={OGC Web Coverage Service Interface Standard}}
\newglossaryentry{wcts}{name=WCTS, description={OGC Web Coordinate Transformation Service Interface Standard}}
\newglossaryentry{wfs}{name=WFS, description={OGC Web Feature Service Interface Standard}}
\newglossaryentry{wg}{name=WG, description={Working Group}}
\newglossaryentry{wmc}{name=WMC, description={OGC Web Map Context Encoding Standard}}
\newglossaryentry{wmo}{name=WMO, description={World Meteorological Organization}}
\newglossaryentry{wms}{name=WMS, description={OGC Web Map Service Interface Standard}}
\newglossaryentry{wmt}{name=WMT, description={Web Mapping Testbed}}
\newglossaryentry{wmts}{name=WMTS, description={OGC Web Map Tile Service Interface Standard}}
\newglossaryentry{wps}{name=WPS, description={OGC Web Processing Service Interface Standard}}
\newglossaryentry{ws}{name=WS, description={Web Services}}
\newglossaryentry{wsc}{name=WSC, description={OGC Web Service Common Interface Standard}}
\newglossaryentry{wsdl}{name=WSDL, description={Web Services Definition Language}}
\newglossaryentry{xima}{name=XIMA, description={XML for Imagery and Map Annotations}}
\newglossaryentry{xml}{name=XML, description={eXtensible Markup Language}}
\newglossaryentry{xslt}{name=XSLT, description={eXtensible Style Sheet Transformation}}


\newglossaryentry{rs}{name=RS, description={Remote Sensing}}
\newglossaryentry{clc}{name=CLC, description={Corine Land Cover}}
\newglossaryentry{modis}{name=MODIS, description={MODerate resolution Imaging Spectroradiometer}}
\newglossaryentry{dhus}{name=DHuS, description={Data Hub Software}}
\newglossaryentry{lucas}{name=LUCAS, description={Land Use and Coverage Area frame Survey}}
\newglossaryentry{lidar}{name=LiDAR, description={Light Detection And Ranging}}
\newglossaryentry{msi}{name=MSI, description={Multi-Spectral Instrument}}
\newglossaryentry{boa}{name=BOA, description={Bottom of Atmosphere}}
\newglossaryentry{toa}{name=TOA, description={Top of Atmosphere}}
\newglossaryentry{olci}{name=OLCI, description={Ocean and Land Colour Instrument}}
\newglossaryentry{gmes}{name=GMES, description={Global Monitoring for Environment and Security}}
\newglossaryentry{esa}{name=ESA, description={European Space Agency}}
\newglossaryentry{ec}{name=EC, description={European Commission}}
\newglossaryentry{snap}{name=SNAP, description={Sentinel Application Platform}}
\newglossaryentry{gpt}{name=GPT, description={Graph Processing Tool}}
\newglossaryentry{wkt}{name=WKT, description={Well Known Text}}
\newglossaryentry{lta}{name=LTA, description={Long Term Archive}}
\newglossaryentry{nor}{name=NoR, description={Network of Resources}}
\newglossaryentry{gdal}{name=GDAL, description={Geospatial Data Abstraction Library}}

\newglossaryentry{tp}{name=TP, description={True Positive}}
\newglossaryentry{fp}{name=FP, description={False Positive}}
\newglossaryentry{fn}{name=FN, description={False Negative}}
\newglossaryentry{tn}{name=TN, description={True Negative}}
\newglossaryentry{iou}{name=IoU, description={Intersection over Union}}
\newglossaryentry{miou}{name=MIoU, description={Mean Intersection over Union}}

\newglossaryentry{utm}{name=UTM, description={Universal Transverse Mercator}}
\newglossaryentry{aoi}{name=AOI, description={Area of Interest}}
\newglossaryentry{mcu}{name=MCU, description={Minimum Cartographic Unit}}
\newglossaryentry{nasa}{name=NASA, description={National Aeronautics and Space Administration}}

\newglossaryentry{api}{name=API, description={Application Programming Interface}}

%% file: sections/abstract.tex
\section*{Abstract} 

In recent years, the geospatial industry has been developing at a steady pace. This growth implies the addition of satellite constellations that produce a copious supply of satellite imagery and other Remote Sensing data on a daily basis. Sometimes, this information, even if in some cases we are referring to publicly available data, it sits unaccounted for due to the sheer size of it. Processing such large amounts of data with the help of human labour or by using traditional automation methods is not always a viable solution from the standpoint of both time and other resources.

Within the present work, we propose an approach for creating a multi-modal and spatio-temporal dataset comprised of publicly available Remote Sensing data and testing for feasibility using state of the art Machine Learning (\gls{ml}) techniques. Precisely, the usage of Convolutional Neural Networks (\gls{cnn}) models that are capable of separating different classes of vegetation that are present in the proposed dataset. Popularity and success of similar methods in the context of Geographical Information Systems (\gls{gis}) and Computer Vision ~(\gls{cv}) more generally indicate that methods alike should be taken in consideration and further analysed and developed.

For the first part, we will start off with an overview of the current problem space and the argument necessity of studying such techniques. Following this, Chapter~\ref{ch:sota} begins by providing extended details regarding the current state of research for~\gls{gis} starting from characteristics of Remote Sensing data to how the acquisition of such data is performed and to additional preprocessings that could be required. Within Chapter~\ref{ch:sota} we also present informations about the inner workings of~\gls{ml} with respect to current heuristics that are used in the literature for solving various~\gls{cv} tasks.

Chapter \ref{ch:approach} showcases the approach taken for developing an appropriate multi-modal spatio-temporal Remote Sensing dataset publicly available data. Some results on experimenting with~\gls{ml} techniques for training Convolutional Neural Networks (\gls{cnn}) on the task of semantic segmentation of vegetation related classes are also presented for validating the usability and relevance of the aforementioned dataset.

Chapter~\ref{ch:experiments} illustrates the results of training such models on the proposed dataset on how they perform on solving the task of semantic segmentation of vegetation.

\pagebreak

\section*{Abstract}

Recent, industria geospațială s-a dezvoltat într-un ritm constant. Această creștere implică și adiția unor constelații de sateliți care produc zilnic o cantitate impresionantă de imagini satelitare dar și a altor tipuri de date specifice teledetecției. Uneori aceste informații, chiar dacă în unele cazuri ne referim la date publice, nu sunt folosite din cauza volumului acestora. Procesarea unor cantități impresionante de date folosind resurse umane sau metode clasice de automatizare nu este întotdeauna o soluție viabilă din punct de vedere al timpului si al resurselor necesare.

În acest document, propunem o abordare pentru crearea unui set de date variat, spațio-temporal compus din date achiziționate prin tehnici de teledetectie, dar și testarea fezabilității acestuia folosind dezvoltari recente din domeniul învățării automate~(\gls{ml}). Mai precis, utilizarea rețelelor neuronale convoluționale~(\gls{cnn}), care sunt capabile de separarea diferitelor clase de vegetație din setul de date propus. Popularitatea și succesul metodelor similare în contextul sistemelor geografice informaționale~(\gls{gis}) și în general în domeniul Computer Vision~(\gls{cv}) indică faptul că metode similare ar trebui luate în considerare și analizate.

In prima parte, vom începe prin a prezenta o viziune de ansamblu a spațiului curent al problematicii si vom argumenta necesitatea studierii tehnicilor de învățare automată. Ulterior, Capitolul~\ref{ch:sota} prezinta detalii extinse legate de starea actuala a literaturii de specialitate atat pentru~\gls{gis}, începând cu unele caracteristici ale datelor preluate prin tehnici de teledetecție până la cum are loc procesul de achiziție al datelor la preprocesări adiționale ce ar putea fi necesare. Capitolul~\ref{ch:sota} include de asemenea informații despre cum funcționează tehnicile de învățare automată vizavi de euristici folosite în prezent pentru rezolvarea a diferite sarcini în domeniul Computer Vision.

Capitolul~\ref{ch:approach} prezintă abordările luate în dezvoltarea unui set de date multi-temporal și spațio-temporal cu informații preluate din teledetecție. Rezultatele experimentelor folosind tehnici de învățare automată prin antrenarea unor Rețele Neuronale Convoluționale (\gls{cnn}) pentru a segmenta semantic imagini satelitare în concordanță cu clase de vegetație pentru a valida setul de date antemenționat.

Capitolul~\ref{ch:experiments} ilustrează performanța modelelor folosind setul de date menționat, aratând eficiența lor în rezolvarea problemei de segmentare semantica a vegetației.

\newpage
\thispagestyle{empty}
\mbox{}


%% file: sections/introduction.tex
\chapter{Introduction}\label{cap:intro}

As of relatively recently, the world has started to become increasingly reliant on technology that implements Artificial Intelligence techniques. From automating labor-intensive tasks, to offering aid in medical diagnoses, to autonomous vehicles, anticipating stock market prices, automatically scaling cloud infrastructures based on predicted demand, natural language processing, data fusion, enhancing image resolution (also named super resolution), image segmentation. The aforementioned are some particular examples of tasks the realm of Artificial Intelligence can offer aid in. There is no doubt about the benefits that a certain subdomain of this field of study, namely Machine Learning (\gls{ml}) is adding to our day-to-day life. We can argue that the increase in popularity for these types of methods is due to the timeframe we live in. Never has the production rate of data been so considerable than now. We are reffering of course, to the Big Data phenomenon. The context in which data is being generated varies. If we are talking about Web 2.0, then we are talking about user generated data, mostly popular nowadays being social media platforms (contents usually include: pictures, videos, audio or text). Some other examples of data sources can be machine generated, such as measurements from sensors, performance and usage logs extracted by using monitoring techniques on cloud and IoT infrastructures.
All of those modern requirements to process large quantities of information have greatly influenced the evolution of Artificial Intelligence. Previous methods of achieving similar tasks, such as expert systems required more focus on defining rules manually (based on the available data) for the system to follow in order to perform reasoning. Nowadays it is nearly impossible to create such rules due to the volume and variety of the data at hand. The goal of Machine Learning algorithms is to automatically generate rules based on the data.




For the purpose of this thesis, we will however focus on the niche of Computer Vision (\gls{cv}). More precisely, on applying Machine Learning to satellite imagery that was collected through remote sensing techniques (detailed in Section~\ref{sec:remote-sensing}). We will use a class of \gls{ml} techniques called supervised learning, which assume that we feed the algorithms with concrete examples (labeled data) of what the input looks like, while also describing the desired output. Therefore, a~\gls{ml} model needs to learn a correlation between the two. Computer Vision itself, is a branch of Artificial Intelligence, and is the discipline dealing with analyzing and interpreting digital images and videos (examples of tasks belonging to \gls{cv}, are detailed in Section~\ref{sec:cv}).



\pagebreak

\section{Remote sensing}\label{sec:remote-sensing}

Generally, by Remote Rensing we understand the process of acquiring information about the physical caracteristics of land surfaces, by utilising methods that do not require direct human contact. As of recently, this industry for both the public and private sectors has seen a steady growth, being an active participant in the big data phenomenon as a data source. Observing the Earth is becoming more important now than ever. With the availability of high quality Remote Sensing data, also comes a research interest that has been growing in the last three decades according to a bibliometric analysis performed on Clarivate's Web of Science database~\cite{duan2020remote}.

\begin{figure}[h!]
  \begin{center}
    \includegraphics[scale=0.5]{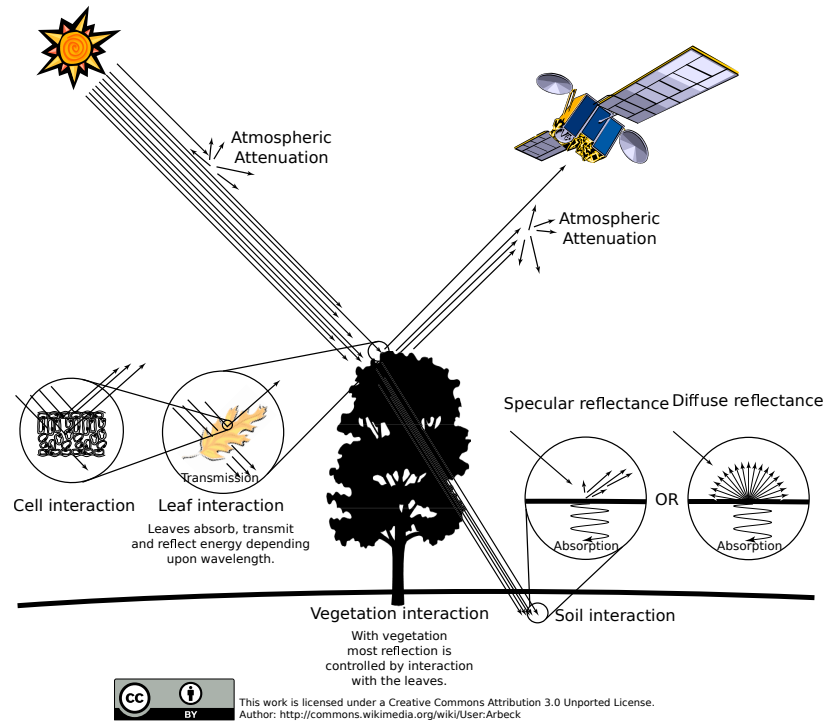}
    \caption{A typical passive Remote Sensing environment}
    \label{fig:remote-sensing-env}
  \end{center}
\end{figure}

There are two categories of remote sensing: either active, or passive, depending on how the acquisition process is done. If the sensors that are used for measurements do not emit radiation, and natural sources such as the Sun energy reflections (visible and infrared wavelenghts) are collected by the sensors, then the process is passive. On the other hand, if sensors are used also for emitting radiation towards the area of interest, whose backscatter\footnote{Reflection of signal back to the source.} will later be collected, then the process is active. Figure~\ref{fig:remote-sensing-env} illustrates the passive process, and Figure~\ref{fig:active-passive-rs} shows the difference between active and passive Remote Sensing. Data acquisition can be performed using multiple instruments: \gls{sar}, \gls{lidar}, \gls{msi}, and \gls{olci} (Section \ref{sec:rs-instruments}) are some of them.

Remote sensing has a wide variety of usecases. Examples include: monitoring land cover or land usage, detecting changes in land coverage, tracking phenomenas with impact on land surface like floodings, movement of glaciers, earthquakes, or forest fires, tracking cloud formations for weather forecasting. The domain also encompasses oceanography, which includes tasks like: scanning the ocean floor, and keeping track of ocean color, turbidity, temperature and other specific properties.

\section{The Copernicus programme}\label{sec:copernicus}

Copernicus\footnote{\url{https://www.esa.int/Applications/Observing_the_Earth/Copernicus/Overview3}}\textsuperscript{,}\footnote{Previously named ``Global Monitoring for Environment and Security'' (\gls{gmes})} is a collaboration between the European Space Agency~(\gls{esa}) and the European Commission~(\gls{ec}). It's aim is to create a powerful Earth Observation (\gls{eo}) programme, starting from launching and controlling of satellites to collecting, processing and distributing the collected data to delivering software products.

By launching the Sentinel family of satellites (see Sections~\ref{sec:sentinel2-description},\ref{sec:sentinel3-olci} for two specific examples), the programme is able to provide easy to access accurate, and high quality remote sensing data. This data ranges from radar and multi-spectral imaging for land monitoring, ocean and land cover data, to atmospheric composition information. Part of the data from the Copernicus programme is being made available through the Data Hub Software\footnote{\url{https://github.com/SentinelDataHub/dhus-distribution}} (\gls{dhus}). The Data Hub Software was used in the scope of this work as well, as it was a primary source for the dataset. Sentinel-2 and Sentinel-3 archives were downloaded in specific, as detailed in Section~\ref{sec:dhus}.

\section{Artificial Intelligence Techniques}~\label{sec:ai-techniques}

\begin{figure}[h!]
  \begin{center}
    \includegraphics[scale=0.7]{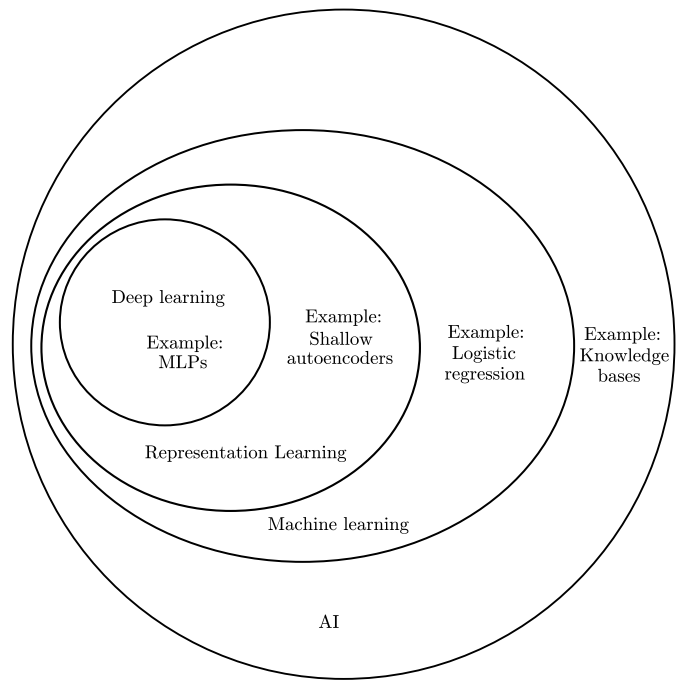}
    \caption{Artificial Intelligence and it's subfields~\cite{goodfellow2016}}
    \label{fig:goodfellow-venn}
  \end{center}
\end{figure}

The Venn diagram illustrated in Figure~\ref{fig:goodfellow-venn} provides us with the context of the current work, with each subset of the Artificial Intelligence superset representing a different branch/subdomain. The methodologies that are to be used in the current work reside in the inner-most set, that of Deep Learning. \gls{mlp} or Multiple Layer Perceptrons are a prominent example of this branch, and they represent the atomic building block that is used to structure each \gls{ann} (Artificial Neural Network). One such class of \gls{ann} that is used extensively in the field of Computer Vision are Convolutional Neural Networks (\gls{cnn}), which are to be described into detail in Section~\ref{sec:cnn}. The outer-most layer (knowledge bases) and some shortcomings when they are used to process larger volumes of data were presented. But what about the other three layers? Should the terms ``Machine Learning'' and ``Deep Learning'' be used interchangeably? Answers to the latter question might vary. There are two main differences between the two. They have to do with the structure of the data, but also with how the learning process is achieved. With Machine Learning, we can for instance provide labeled input images. The model is trained in order to progressively get better at the task of classifying images based on labels (correlation of input-output). It can only take into account the global context of the image we are providing in order to determine the features that dictate what label it should assign to it. Deep learning assumes there exists an hierarchy of layers, each of it being able to determine local features within the image that will build up the reasoning behind choosing the class an image belongs to. Furthermore, they can be extended with the help of Softmax (Section~\ref{sec:softmax}) into providing a probability vector regarding the beloging of a particular input to a class, or to what the problem at hand might be. This is in contrast to another type of~\gls{cnn}, namely Fully Convolutional Networks (\gls{fcn}) where the last layer is replaced by a convolution one instead of Softmax or other classifiers. There is an importance of how data is represented and given to a Machine Learning model. Sometimes there is the possibility to hand-pick or craft\footnote{This is also known as ``feature engineering''.} the features which are to be fed to the model. But in the case where this is not possible and representation becomes more complex such as when dealing with comprehending images or videos or processing natural language, Representation Learning is a viable choice. Representation Learning provides ways of automatically learning feature representations. Common ways of learning such representations are autoencoders (where representation is learned separately from labeled data\footnote{Therefore, autoencoders are categorised as unsupervised learning.}). Approaches based on encoder-decoder topologies (some examples are discussed within Chapter~\ref{ch:sota}) use said encoders for mapping inputs to feature representations, while also using them in the decoder for taking informed decisions.



\section{Motivation}\label{sec:motivation}

This section provides an answer to an important question: \textit{``What do we have to gain from using Machine Learning with this kind of data?''}. Presuming the proposed task is being executed successfully (i.e.\ the models being able yield a satisfactory performance\footnote{In comparison to similar results of methods that are in the same class of problems} in respect to semantic segmentation) there are numerous benefits within the realm of Remote Sensing and Geographic Information Systems (\gls{gis}) in particular. While also having an impact on the human decision-making process regarding organizational measures in general.

Focusing on the task of semantic segmentation (Section~\ref{sec:cv}) in Remote Sensing imagery, the end result, the prediction masks (or segmented images) provide an overview of land usage/land cover. In this work, the prediction masks are displaying the composition and diversification of vegetation in the images given to the model.

Before going further, there are some explanations to be done about the two terms mentioned in previous paragraphs. Altough land use and land cover are sometimes used interchangeably, they reffer to distinct issues. The first term's meaning is that of how parts of the Earth's surface are being used by the population, and it is related to socioeconomic aspects as well. The latter's is strictly reffering to the physical composition of the surface of Earth. In both cases, the end product, usually called an inventory, is comprised of a set of labels, which are associated to real-world surfaces within the area of interest in which the study was conducted on.


The current proposal falls into the category of land coverage. Typically land cover information is collected by one of two means: either through land surveys or through Remote Sensing techniques. There are two examples of such inventories, which are to be discussed in Chapter~\ref{ch:sota}. Namely, we are talking about the LUCAS\cite{ballin2018redesign} dataset that was created through land surveys conducted by an initiative of Eurostat. On the other hand Corine Land Cover (\gls{clc})\cite{bossard2000corine} is a land cover inventory created with the availability of Remote Sensing imagery (by using multiple satellites such as Landsat, SPOT and since 2018 Sentinel-2 and Landsat-8). The data was later inspected, and the inventory was generated by using semi-automated~\gls{gis} methods.

Typically, land cover and Remote Sensing in general are considered to be sub-domains of geography. Previous approaches to process remote sensing data were performed either manually by cartographers (e.g.~in the case of CLC~\cite{bossard2000corine}), or with the help of morphological transformations~\cite{tsoeleng2020comparison, plaza2005dimensionality}. Morphological transformations are a set of mathematic operations, which are usually applied to images in order to perform various processings. While these techniques yield satisfiable performance, they are limited in terms of the variety of tasks they can perform. It is also common for morphological transforms to be used in conjunction with artificial intelligence~\cite{benediktsson2003classification} in order to preprocess or postprocess data, thus ehancing the performance of models.

Once obtained, information about land coverage can be of use in multiple different applications. Most of which are concerns that qualified institutions are expected to work with as part of their expertise. This list includes:

\begin{itemize}
\item Environmental planning
\item Monitoring the effects of climate change
\item Disaster management and risk assessment
\item Environmental predictions
\item Management and preservation of wildlife
\end{itemize}


All of the aforementioned applications are dependent on information concerning the physical coverage of a given area of interest. Appendix~\ref{sec:landcover-info} provides additional details about the land cover on the territory of Romania, providing further motivation for studying changes land cover for vegetation. The current work proposes a method for extracting land cover inventories automatically, with interest in vegetation coverage by using remote sensing data with the help of modern artificial intelligence.

\pagebreak




\paragraph*{A short description of the problem}

\mbox{}\\

The current work contains some proposals for developing Machine Learning models that have the capability to differentiate between different types of vegetation using a new multi-modal and spatio-temporal dataset comprised of Remote Sensing data.

There exist limitations regarding the current state of the art in both Artificial Intelligence, and with the data that is to be processed. For example: even if desired, it is not possible to obtain a finer granularity over the number of classes that we predict, if our dataset does not contain concerete examples for more classes (i.e.\ the classes that we train our model with, are also the ones that it outputs in predictions).

\begin{figure}[h!]
  \begin{center}
    \includegraphics{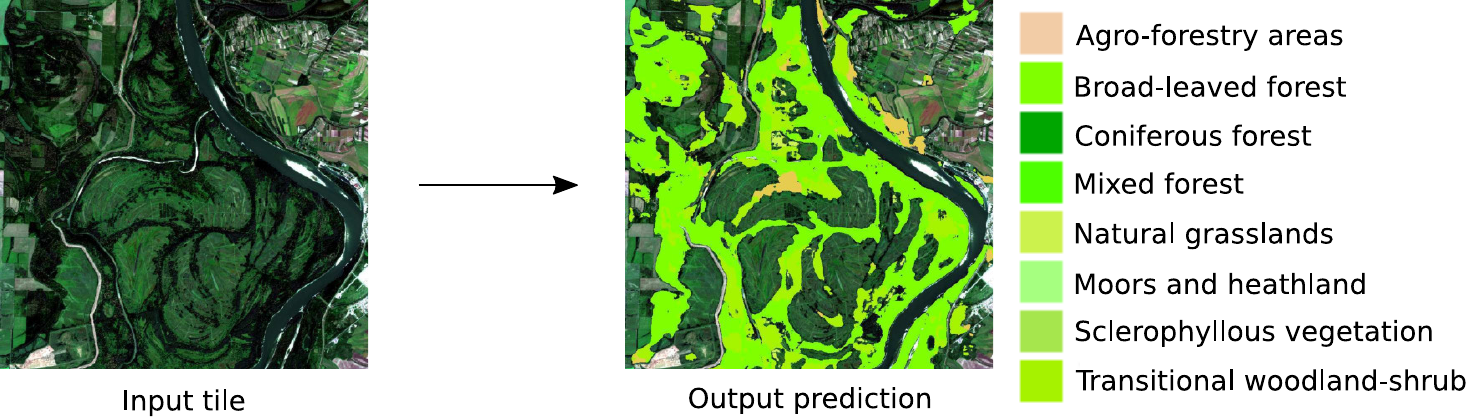}
    \captionsetup{justification=centering}
    \caption{Example of semantic segmentation of forestry in Sentinel-2 RGB data\\\tiny{Images extracted from LeafS project developed for FOSS4G 2019 EO Data Challange}}
    \label{fig:segmentation-forestry-example}
  \end{center}
\end{figure}

The simplest example of what the goal of the current work is illustrated within Figure~\ref{fig:segmentation-forestry-example}. Based on a given input, in this context a satellite image, through using Machine Learning techniques, we are to be able to generate a prediction mask indicating the types of vegetation that exist in the image and where they reside.

\paragraph*{Objectives of the current work}

\mbox{}\\


\begin{itemize}
\item{Creating a multi-modal (Section~\ref{sec:multi-modal}) dataset covering the area of Romania.}
\item{Training of various Machine Learning models based on convolutional neural network topologies (Sections~\ref{sec:segnet},~\ref{sec:unet},~\ref{sec:resnet}) using Hugin (Section~\ref{sec:hugin}) on the task of semantic segmentation for the vegetation classes from \gls{clc} using the aforementioned multi-modal dataset.}
\item{Analyzing the results and comparing the performance yielded by these models.}
\end{itemize}


%% file: sections/sota.tex
\chapter{State of the Art}\label{ch:sota}

The current chapter provides an overview of current literature in both Artificial Intelligence and Remote Sensing, focusing on techniques that are used in Chapter~\ref{ch:approach}.

\section{Data acquisition techniques}\label{sec:rs-instruments} 

As briefly mentioned in Section~\ref{sec:remote-sensing}, there are various instruments on board of satellites. For the purpose of the current work, we want to focus on two of them: ~\gls{msi} (Multi-Spectral Imagery) provided with Sentinel-2 and~\gls{olci} (Ocean and Land Colour Instrument) from Sentinel-3. These instruments provide information about land coverage that can be useful when detecting vegetation in Remote Sensing data.

The Sentinel-2~\gls{msi}\footnote{\href{https://sentinels.copernicus.eu/web/sentinel/user-guides/sentinel-2-msi/overview}{\nolinkurl{sentinels.copernicus.eu/web/sentinel/user-guides/sentinel-2-msi/overview}}} data contains 13 spectral bands at different spatial resolutions (4 bands at 10m resolution, 6 at 20m and 3 bands at 60m). Various types of vegetation can be identified based on their signature, that is the spectral response collected as backscatter by the satellite's instruments. Data is distributed via the ~\gls{dhus} in different processing levels. In the case of Sentinel-2~\gls{msi} there are three processing levels, but only two are publicly available. Level-0, Level-1 and Level-2 each including different stages of processing. Level-2 data is the most ``complete'' in terms of processing, consisting of ~\gls{boa} (Bottom of Atmosphere) reflectance images that are georeferenced in the \gls{utm} grid projection. The granularity of a Sentinel-2~\gls{msi} Level2-A scene (or tile) is $100$x$100km^2$. Sen2cor~\cite{main2017sen2cor, baetens2019validation} is further used in order to process Sentinel-2 archives and also detect clouds, cloud shadows and snow. This method is helpful to take into consideration when using Machine Learning, as portions of images that are occluded by clouds can interfere with training models.


The Sentinel-3 mission also provides data from the ~\gls{olci}\footnote{\href{https://sentinels.copernicus.eu/web/sentinel/user-guides/sentinel-3-olci/overview}{\nolinkurl{sentinels.copernicus.eu/web/sentinel/user-guides/sentinel-3-olci/overview}}} instrument. At a 300m spatial resolution, a product is composed of data from 21 spectral bands, quality control flags, tie points and information regarding georeferencing. The 21 bands encompases different functions raging from chlorophyll absortion, sediment loading, chlorophyll fluorescence levels, water vapour absorption levels to aerosol corrections and atmospheric corrections. This information makes Sentinel-3 ~\gls{olci} a good instrument for identifying signatures of vegetation in land coverage.


\section{Remote Sensing centric datasets}\label{sec:rs-datasets}

In the context of Remote Sensing, several datasets are present in the literature and are publicly available. Most of them have been created with the purpose of helping building automated methods that give aid in identifying the land coverage or usage.


Two most common ways of storing \gls{gis} data are by using either vector or raster formats. With vector formats, information is stored based on drawn shapes (such as points, lines, polygons) using geographical coordinates as a spatial reference. Vector formats typically store discrete values. Raster is a matrix like structure that contains features usually stored as continuous values, with a fixed spatial resolution. Vectorial dataset examples include \gls{clc} and \gls{lucas}, while BigEarthNet, \gls{modis}, EuroSAT and SpaceNet contain raster data. Rasterization is the process of converting vector format data into raster data (as exemplified within Section~\ref{sec:rasterization}).

\subsection{SpaceNet Roads and SpaceNet Buildings}\label{sec:spacenet}

In the context of a competition, CosmiQ Works in collaboration with Amazon have published the SpaceNet~\cite{van2018spacenet} dataset. The objective in this competition is to create~\gls{ml} models that yield a good performance for binary semantic segmentation of road networks or buildings, or for monitoring urban development over time. The dataset contains more than 10.500km of road structures and 811,982 polygons marked as buildings. SpaceNet7 includes 24 satellite imagery mosaics representing a timeframe of 24 months covering over $40.000km^2$ representing urban development.


The dataset is composed of RGB and multi-spectral WorldView2 and WorldView3 imagery at a spatial resolution of 30, respectively 50 centimeter. Besides the input images, ground truth labels are provided as GeoJSON and they have been extracted from \href{https://openstreetmap.org}{OpenStreetMaps} and manually checked to correspond to real road networks or building structures. The competition is developed to increase in complexity with each iteration by adding more requirements, such as estimating the time it takes to traverse a predicted road network or by adding off-nadir\footnote{By nadirness, we understand the various angles of the satellite in relation to the target at the moment of acquiring the images. Nadir ($<25^{\circ}$), off-Nadir ($>25^{\circ}, <40^{\circ}$), vert off-Nadir ($>40^{\circ}$).} and very off-nadir images.




\subsection{BigEarthNet}\label{sec:bigearthnet}

BigEarthNet~\cite{sumbul2019bigearthnet}, a recently published remote sensing image benchmark dataset, is composed of 590,326 tiles which were extracted from 125 Sentinel-2~\gls{msi} archives. The tiles have different spatial resolutions (126x126 for the 10m bands, 60x60 for 20m bands, and 20x20 for the 60m bands). This dataset is specifically designed for training and comparing the accuracy of deep learning models (benchmarking). The sensing period of the data ranges from between June 2017 and May 2018, and the Area of Interest (\gls{aoi}) covers 10 European regions (namely Austria, Belgium, Finland, Ireland, Kosovo, Lithuania, Luxembourg, Portugal, Serbia and Switzerland).

Each image is annotated with one or more labels from the Corine Land Cover inventory (Section~\ref{sec:clc}). The absence of segmentation masks and small tile sizes make BigEarthNet better suited for training ~\gls{ml} models in order to solve the task of image recognition (or classification) rather than for solving semantic segmentation.


\subsection{Corine Land Cover (\gls{clc})}\label{sec:clc}

Corine Land Cover\footnote{\url{https://land.copernicus.eu/pan-european/corine-land-cover}} is a popular land cover/land use inventory created within the context of ESA's Copernicus Land Monitoring Service. It mainly consists of a hand-mapped (based on raster interpretation) package of polygons belonging to 44 different land cover classes for countries that are either part of the European Union, or who volunteer to provide data. There are some countries have used semi-automated methods to create their inventory (Germany, Ireland, Spain, Portugal).

The inventory exists since 1990, and the latest remapping has been performed in 2018. A variety of different satellite data has been used as guides for it's creation: 1990 - Landsat-5 (50m spacial resolution), 2000 - Landsat-7(25m),  2006 - SPOT-4/5 (25m), 2012 - IRS P6 LISS III and RapidEye (25m), and the latest, 2018 - Sentinel-2 (10m). Changes in land coverage are also distributed separately in \gls{clc}. The minimum cartographic unit (\gls{mcu}) in \gls{clc} is 25ha, and the accuracy of the mapping is better than 100m. Data within the inventory is stored as a hierarchy on three levels based on grouping of classes, as mentioned in the nomenclature~\cite{feranec2016corine}.

The extents of this inventory (39 countries partaking in the effort by 2018) as well as the fact that the inventory is performed and published periodically and changes in land cover are marked separately together with the variety of existent classes make Corine Land Cover a viable choice for using it as ground truth data when investigating land coverage, usage or change detection on the European continent.

\begin{figure}[h!]
  \begin{center}
    \includegraphics[scale=0.15]{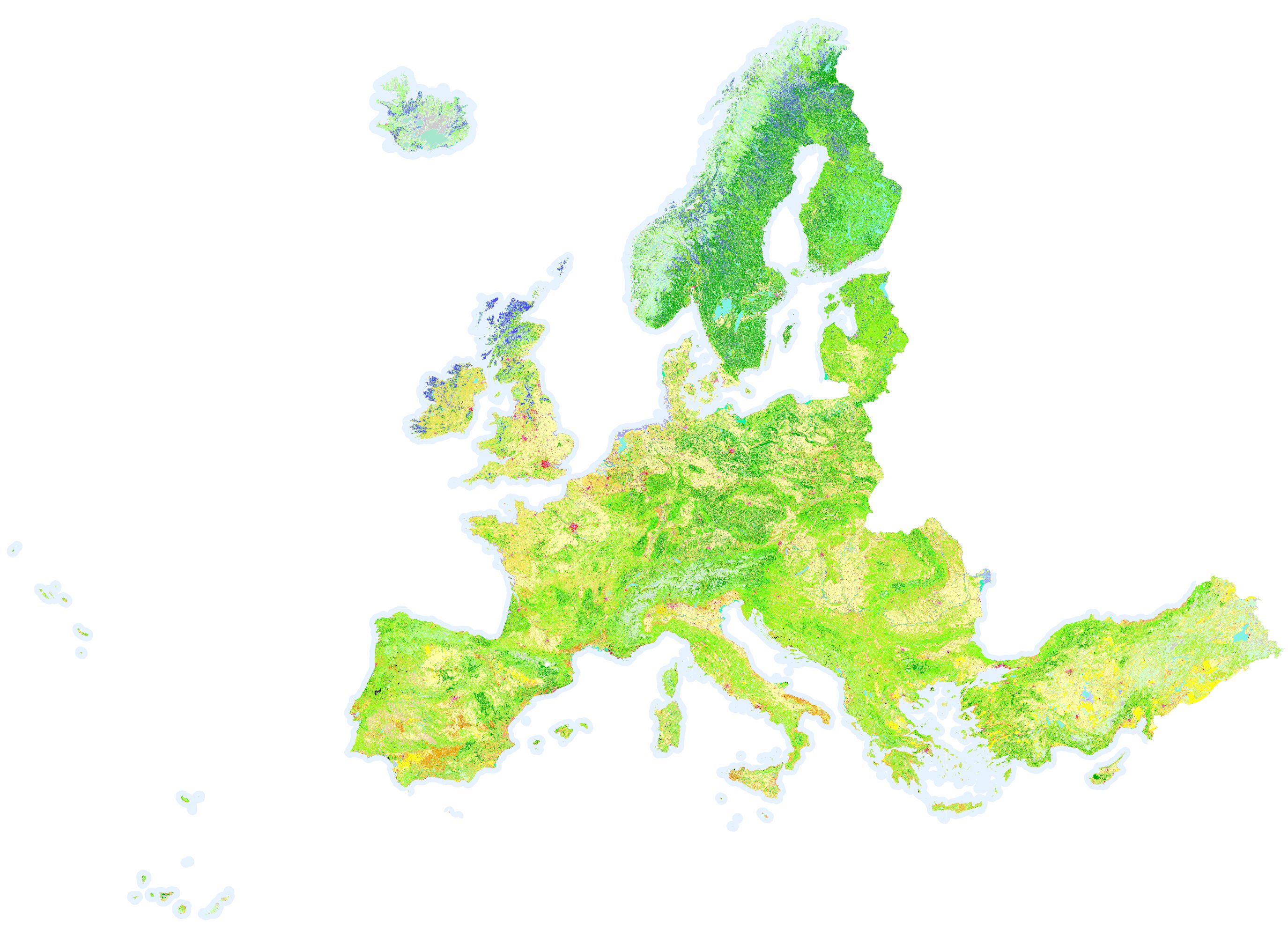}
    \caption{Extents of the Corine Land Cover inventory}
    \label{fig:clc}
  \end{center}
\end{figure}

\pagebreak

\subsection{\gls{modis} Land Cover}\label{sec:modislc}

\begin{figure}[h!]
  \begin{center}
    \includegraphics[scale=0.9]{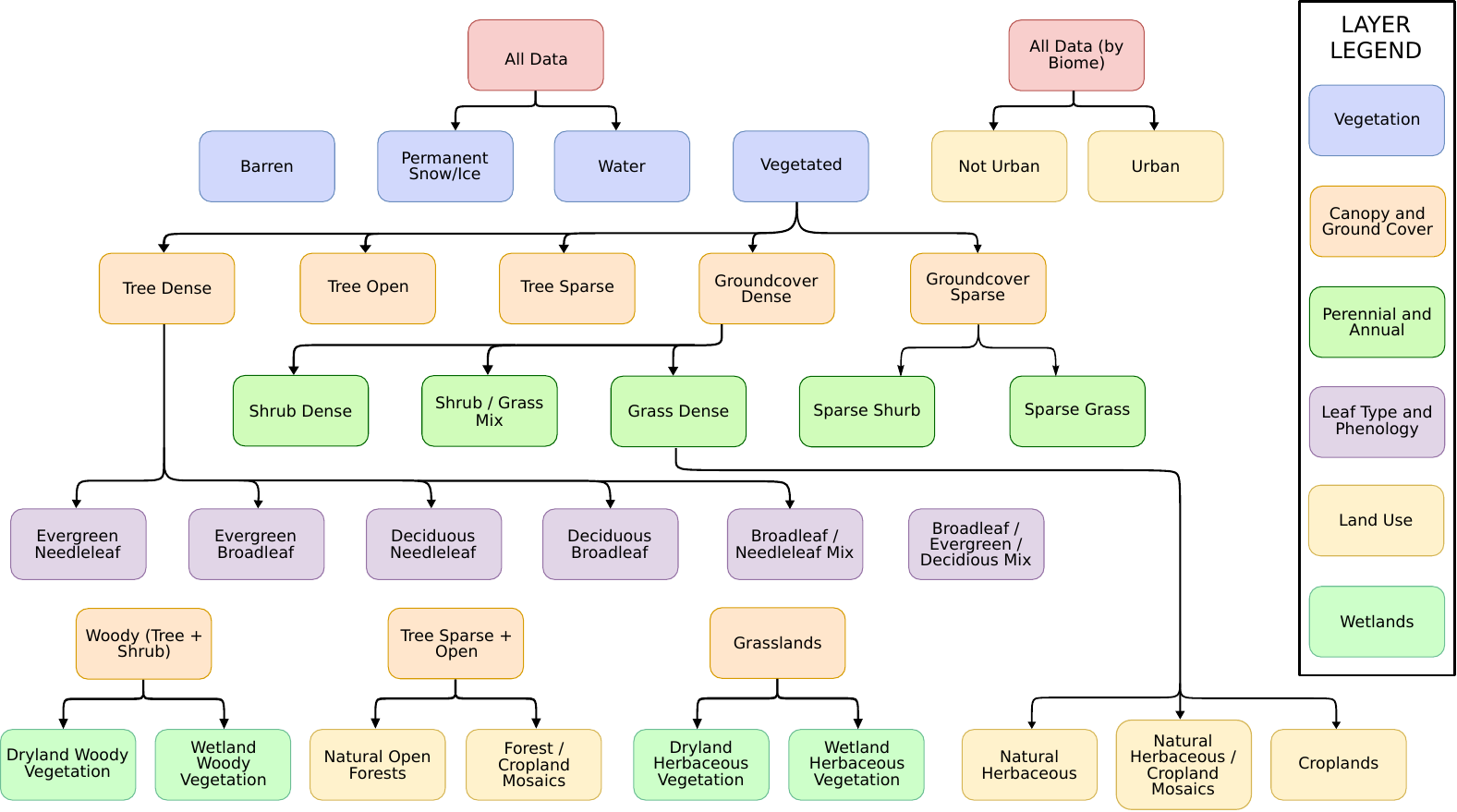}
    \caption{MODIS Land Cover classes hierarchy~\cite{sulla2019mcd12q1}}\label{fig:mcdschemes}
  \end{center}
\end{figure}

Similarly to ESA's effort in the field of land monitoring and land cover, \gls{nasa} also releases the Moderate Resolution Imaging Spectroradiometer (\gls{modis}) Land Cover product~\cite{sulla2019mcd12q1} (or MCD12Q1). Currently at it's 6th version, this inventory has been released annually since 2001. It is composed of 13 datasets, which spread hierarchically on 5 different classification schemes (they are shown in Figure \ref{fig:mcdschemes}). The datasets also include layers for quality assurance and a binary land-water mask.

The \gls{modis} MCD12Q1 land cover inventory is created with the help of classification models~\cite{friedl2010modis} using reflectance data acquired worldwide from the \gls{modis} instrument. The inventory that is derived from these models has a spatial resolution of 500m.

\subsection{LUCAS}\label{sec:lucas}

\gls{lucas}~\cite{ballin2018redesign} or Land Use and Coverage Area frame Survey performed by Eurostat\footnote{\url{https://ec.europa.eu/eurostat/home}} for mapping changes in land coverage and land usage over time for the entire territory of the European Union. This survey has been taking place regularly at a 3 year interval since 2006 (altough the ~\gls{lucas} initiative has been in existence as of 2001).

The land cover and use data in the ~\gls{lucas} inventory is collected manually. This means that surveyors have to physically assess the landscape within the ~\gls{aoi} (also called \textit{``in situ''}) in order to determine the coverage and usage and even take soil samples to be further examined\footnote{\url{https://ec.europa.eu/eurostat/web/lucas/overview}}. Besides having to determine land cover/use, the surveyors also take photographs in the 4 cardinal directions. Data acquired by the surveyors is photo-interpreted and land cover/use in any given point is classified from a catalog of 8 categories, 29 classes and a total of 76 subclasses. Data was collected at points at a distance of 2 kilometers to eachother and the survey contains a total of around 1.1 million such points in which observations were made~\cite{ballin2018redesign}.

\begin{figure}
  \begin{center}
    \includegraphics[scale=0.2]{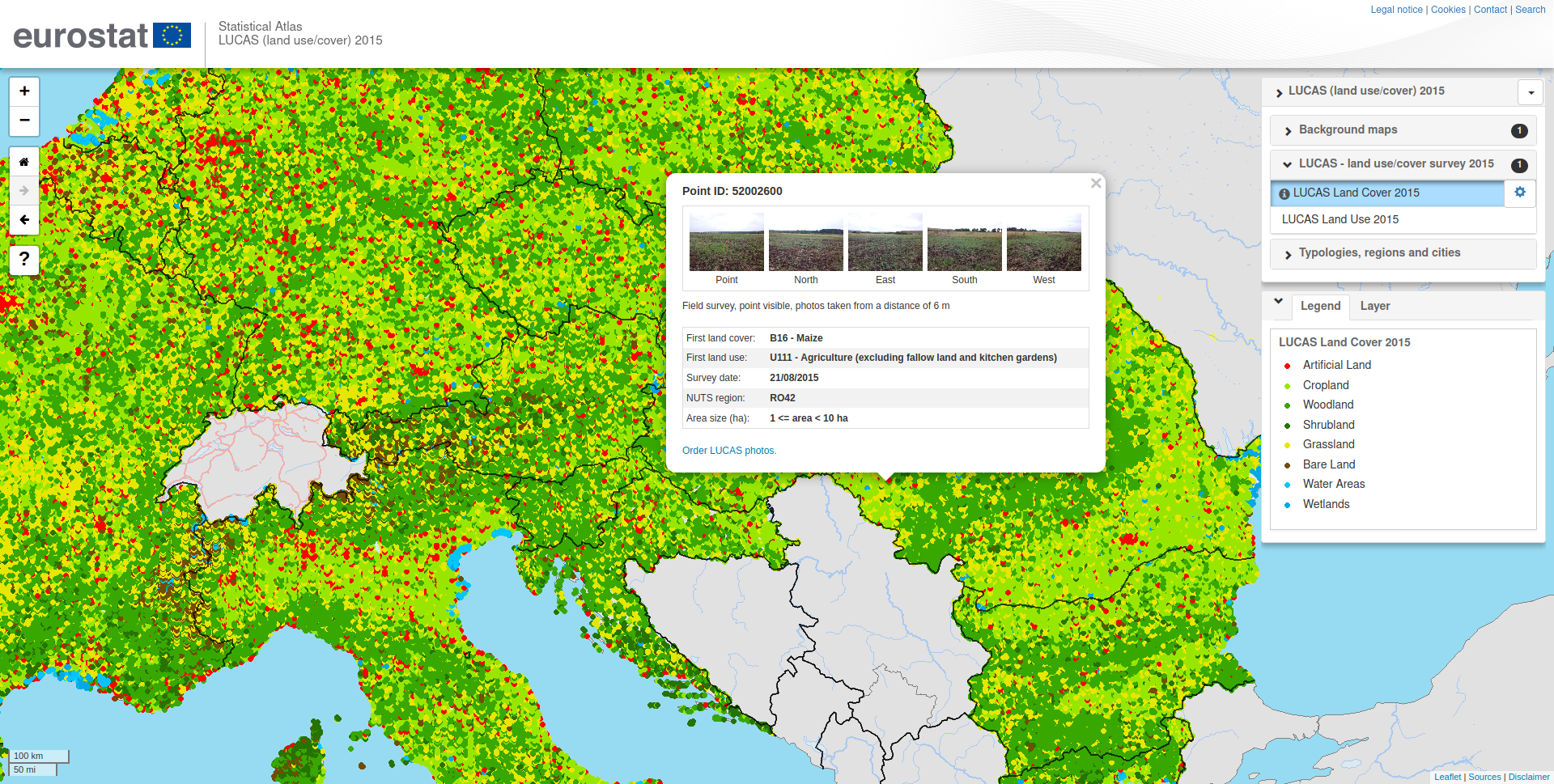}
    \caption{Screenshot of LUCAS Statistical Atlas}~\label{fig:lucasportal}
  \end{center}
\end{figure}

Access to the data aquired from the \gls{lucas} survey is achieved through the web portal\footnote{\href{https://ec.europa.eu/eurostat/statistical-atlas/gis/viewer/?config=LUCAS-2018-HiRes.json}{\nolinkurl{https://ec.europa.eu/eurostat/statistical-atlas}}} (Figure~\ref{fig:lucasportal}). The data that's being made available consists of aggregated statistical tables and individual points data (images, classes of land cover and use).

Overall, with the fact that the data is collected manually presents an advantage in the context of building ~\gls{ml} models, since the ground truth data is highly accurate.

\subsection{EuroSAT}~\label{sec:eurosat}

EuroSAT~\cite{helber2019eurosat} is a benchmarking dataset designed for identifying land usage and land coverage. Composed out of 27.000 Sentinel-2~\gls{msi} imagery that have been labeled using 10 land cover classes. Satellite imagery for 34 countries located in Europe (that are associated with the European Urban Atlas\footnote{\url{https://land.copernicus.eu/local/urban-atlas}}) that contain a small cloud coverage have been hand picked. The size of an image patch found in EuroSAT is 64x64 pixels and the labels have been extracted from the European Urban Atlas.

For statistical relevance, EuroSAT data was picked uniformly throughout the year of 2019 and each land cover class has about 2000-3000 samples spread across cities from the 34 countries. Benchmarking and proof of the relevance was performed using \gls{ml} techniques~\cite{helber2019eurosat}, namely by using \gls{cnn} for listing classification accuracy. More precisely, EuroSAT~\cite{helber2019eurosat} has been validated by using the ResNet-50~\cite{he2016deep} and GoogLeNet~\cite{szegedy2015going} topologies for classifying land coverage on different splits of data.

Due to the small size of patches (64x64 pixels) and the absence of segmentation masks, EuroSAT~\cite{helber2019eurosat} is similar to BigEarthNet~\cite{sumbul2019bigearthnet}. This means it also has a good potential for using it as a benchmarking dataset to be used for solving problems such as classification of land cover classes, but it is not ideal for semantic segmentation.

\vfill



\pagebreak





\section{Machine Learning in the context of Computer Vision}

The primary goal in Computer Vision is to be able to automatically interpret and understand image and video content mimicking as closely as possible the human process of interpreting vision. The usage of Artificial Intelligence techniques such as Machine Learning can provide aid to in understanding this type of content.

Even if Machine Learning, especially applied in the field of computer vision is a hot research topic, it is a fairly mature field. Nowadays, Approaches regarding Deep Learning techniques, especially Convolutional Neural Networks such as the ones shown in Sections~\ref{sec:segnet},~\ref{sec:unet}, are popular amongst research within the field of \gls{cv}.

Most tasks in the field of Computer Vision are solved with the help of supervised learning methods such as Convolutional Neural Networks proven to yield satisfactory results. Using supervised Machine Learning techniques to solve Computer Vision tasks such as the ones described in Section~\ref{sec:cv} also comes with difficulties. These dificulties are discussed within Section~\ref{sec:ccv}. Since research in the field of~\gls{cv} is popular, methods for mitigating those difficulties are also known to exist.

With the appearance of datasets specialized to provide aid in~\gls{cv}, some examples including: ImageNet~\cite{deng2009imagenet}, CIFAR~\cite{krizhevsky2009learning} or MS-COCO~\cite{lin2014microsoft}, solving tasks (Section~\ref{sec:cv}) using supervised~\gls{ml} approaches, especially by using~\gls{cnn} has been widespread. Altough there are numerous specialized datasets that are ready to provide aid in training automated techniques to perform on Computer Vision tasks, there is numerous content being generated on a daily basis from Web2.0, satellites, and generaly simple cameras, most of this data is not labeled in a way to permit supervised Machine Learning. Development of datasets is an important part of contributing to both the fields of Computer Vision and Machine Learning.

Some datasets are distributed in centralized places, such as on the Kaggle\footnote{\url{https://www.kaggle.com/}} platform which is destined to create competitions for developing automated methods for solving tasks. Kaggle hosts numerous~\gls{cv} competitions which have labeled datasets for performing supervised learning.

Applications of Machine Learning in the context of Computer Vision varies to a multitude of domains, ranging from building autonomous vehicles (which include detection of traffic signs, road condition monitorization, traffic monitoring, pedestrian detection), providing aid in medical diagnosis by interpreting images taken from different medical aperture, facial detection and recognition, monitoring of crops, to changes in vegetation and urban development.

\pagebreak

\subsection{Image Classification, Localization, Object Detection, Instance Segmentation and Semantic Segmentation}\label{sec:cv}

\begin{figure}[h!]
  \begin{center}
    \includegraphics[scale=0.2]{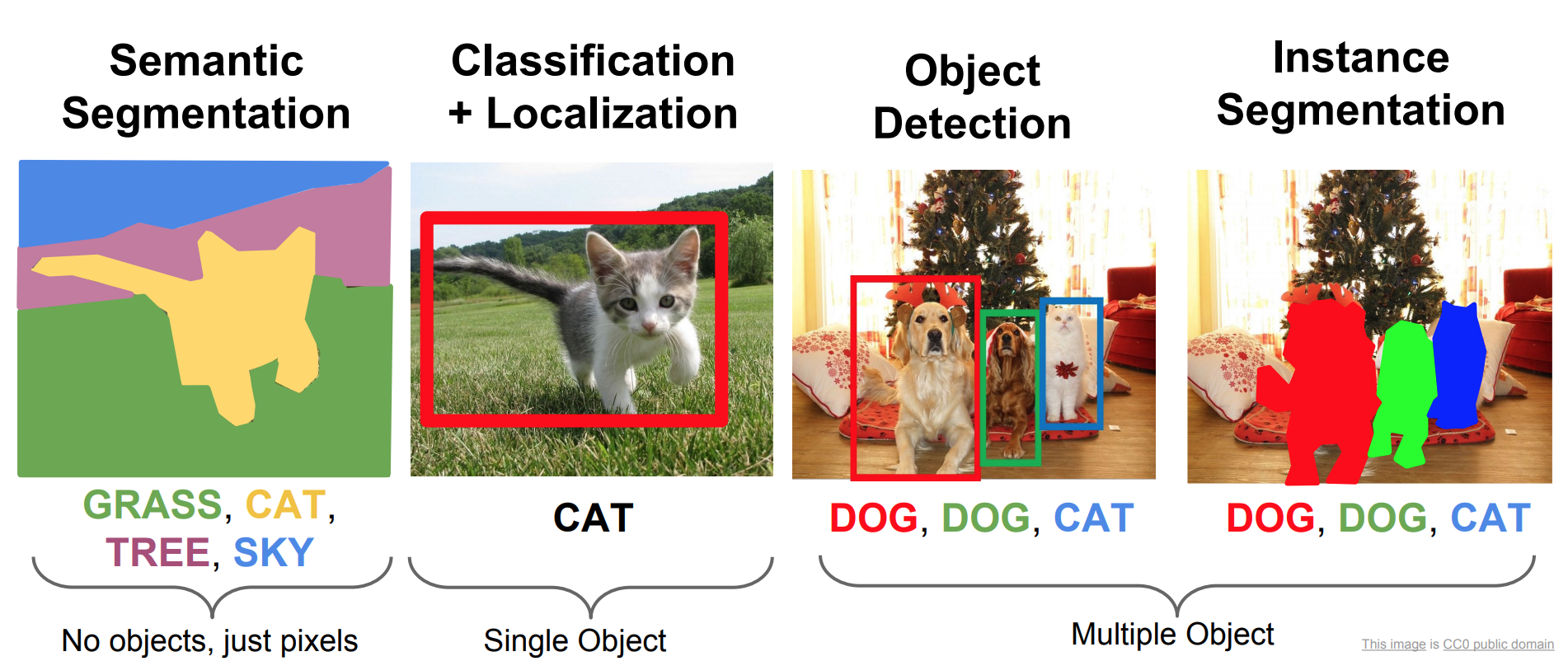}
    \caption{Comparison of the four common tasks in Computer Vision~\cite{karpathy2017cs231n}}
    \label{fig:cvtasks}
  \end{center}
\end{figure}

There are multiple different problematics to be solved within the field of Computer Vision, and also, some open problems as well. The most popular tasks are illustrated within Figure~\ref{fig:cvtasks}. Due to the fact that the goal of the current work (i.e.~Semantic Segmentation) is built upon knowledge from the other tasks, we will provide an explanation for what each of those suppose and the key differences between them. Altough these task might seem trivial to solve at first, the open problems such as those listed within Section~\ref{sec:ccv}, or the existence of adversarial patches~\cite{46561} make solving Computer Vision tasks significantly more difficult.

\textbf{Classification} is when an image is assigned a single label\footnote{From a given, fixed set of labels that we train a model with.} representing the most prominent object contained in an imput image. As illustrated in Figure~\ref{fig:cvtasks}. A couple examples of well-known datasets which are specifically constructed with tasks such as classification in mind are: CIFAR~\cite{krizhevsky2009learning}, ImageNet~\cite{deng2009imagenet} or MS-COCO~\cite{lin2014microsoft}.

\noindent \textbf{Classification with localization} in addition to simple classification, localization also encompasses the drawing of a bounding box over a section of the original input image. Therefore strictly encasing the one object that was classified by the model.

\noindent \textbf{Object detection} is about classifying multiple objects from a single input image. This detection is done in the style of localization. The output is then comprised of multiple bounding boxes and labels corresponding to the detected objects.

\noindent \textbf{Instance segmentation} is similar to object detection, but with some additions. Instead of having bounding boxes, the output now is a pixel mask over the image. Most important, it adds the ability to distinguish between objects of the same class.

\noindent \textbf{Semantic segmentation} refers to partitioning a given input image into regions of pixels which cover each of the different classes that compose the image. The output in semantic segmentation is, again, a mask of pixels with the size being equal to the one of input image. This task is what we want to adress in respect to Remote Sensing data in order to accurately classify different classes of vegetation (available within the Corine Land Cover inventory - see Section~\ref{sec:clc}). Furthermore, the methods used to approach the problem are described within Chapter~\ref{ch:approach}.

\subsubsection{Challenges in Computer Vision}\label{sec:ccv}

There are numerous problems that interfere with solving the problems that we're presented in Section~\ref{sec:cv}. All of them have to do with the nature of the data that we have available. These problems are also encountered in Remote Sensing imagery.

\bigskip

Some open problems within Computer Vision include:

\begin{itemize}
\item \textbf{Viewpoint variation} is when an image of an object is acquired at different orientations in relation to the device that acquired the image.
\item \textbf{Scale variation} happens when instances appear at varied sizes not due to the extents of the image, but to their real-world size.
\item \textbf{Deformation} where real-world objects appear distorted in the images.
\item \textbf{Occlusion} happens when the object of interest is either partially or almost totally covered by another object.
\item \textbf{Illumination variation} are changes in the lightning conditions of images. They can also create issues when there are significant changes of it throughout examples within a dataset.
\item \textbf{Background clutter} sometimes, it could be difficult even for the human eye to distinguish objects within a cluttered image.
\item \textbf{Intra-class variations} reffer to the dissimilarities between instances of the same class. Instances that are part of the same class might look very different.
\item \textbf{Adversarial patches}~\cite{46561} present an interesting approach: generating patches from an image (usually, an image containing a random object) with some specific means of distorsion and applying them to an image that is going to be fed to a Machine Learning model to predict. The goal here is to purposefully trick the model into miss-classifying the object found in the image.  
\end{itemize}

\begin{figure}[H]
  \begin{center}
    \includegraphics[scale=0.43]{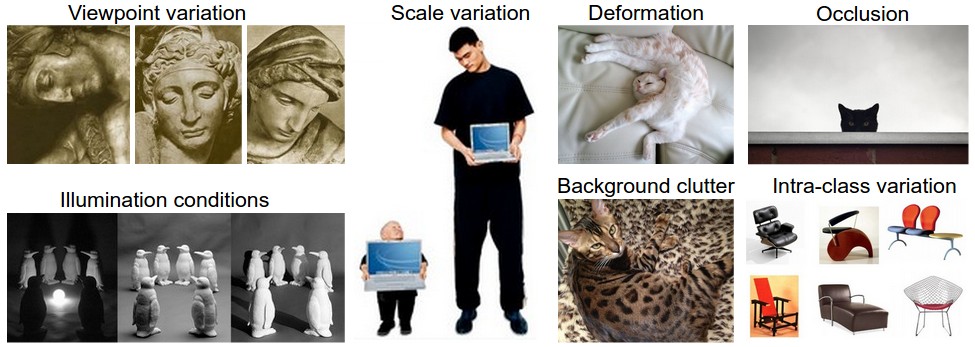}
    \caption{Challenges in Computer Vision~\cite{karpathy2017cs231n}}
    \label{fig:cv-challanges}
  \end{center}
\end{figure}

With the exception of adversarial patches, these are shown within Figure \ref{fig:cv-challanges}.

The challenge that is most encountered when working with Earth Observation data is occlusion. Mainly, this problem appears especially when working with RGB imagery and cloud formations are the objects that are occluding data. There exist several approaches which help mitigate the problem such as creating cloud masks~\cite{baetens2019validation} in order to take occluded data into account when training models. These techniques have been extracted from the literature and discussed are within Section~\ref{sec:cloudmasks}.


\section{Convolutional Neural Networks}\label{sec:cnn}

\begin{figure}[h!]
  \begin{center}
    \includegraphics[scale=1]{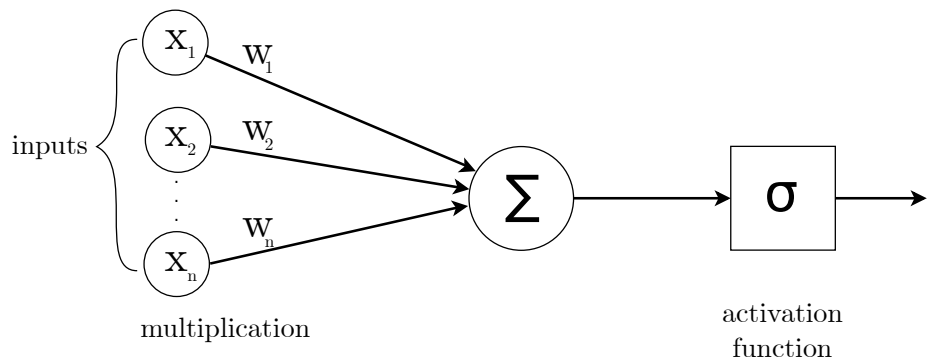}
    \caption{Example showing a perceptron, the principal building block of CNN's.}
    \label{fig:perceptron}
  \end{center}
\end{figure}

Convolutional Neural Networks (\gls{cnn}) are a class of deep learning networks, that are widely used on matrix-like data. They have been introduced to the literature ever since 1989 with the purpose of classifying hand-written zip-codes~\cite{lecun1989}. Convolutional Neural Networks are organized in a multiple layer perceptron (Figure \ref{fig:perceptron}) fashion (i.e.~there exists one input layer for the data, followed by multiple hidden layers containing different operations, which are described into detail in Section~\ref{sec:ml-op}, and finally, at the end of the topology there exists an output layer).

These perceptrons are the virtual equivalents of human neurons, and in a typical neural network, they are aggregated to pass their output as input for the next layer of perceptrons. The basic functioning principle of multi-layer perceptrons is that, for the first layer of the network (i.e.~the input layer) the input vector $[x_1, x_2, \ldots, x_n]$ corresponds to the values of the input data (typically the values of an image from the dataset). These values are multiplied by weights $[w_1, w_2, \ldots, w_n]$ which are modified by the training algorithm, called backpropagation (Section~\ref{sec:backpropagation}). The weighted sum is passed to an activation function (Section~\ref{sec:activation-funcs}) that decides if the perceptron is or not to be activated (i.e~it contributes to the reasoning process).

\subsection{Backpropagation: How are Neural Networks trained?}\label{sec:backpropagation}

Backpropagation is the algorithm used for training a certain class of \gls{ann}, namely feedforward neural networks (meaning that the connection between perceptrons in the network are unidirectional and acyclic). Since an \gls{ann} is comprised of multiple layers of perceptrons, we have multiple weights between these layers. We also recieve an input vector x, and we want to create an association between the input x and the output y. The goal of backpropagation is to understand how to adjust the biases and the weights between layers in order to minimize the error of a model (calculated by an error function). The error is therefore calculated as a difference between what a model predicts ($y_i$) and what the example in the training dataset actually is ($\hat{y}_i$).


\subsubsection*{Underfitting and Overfitting}\label{sec:under-overfitting}


In the context of \gls{ml}, whenever a model is not able to correlate input and output, the performance will be terrible. Underfitting is encountered when there are few training samples and it typically presents with a high rate of error for both training and testing. In contrast to this, overfitting occurs when the model cannot generalize on the task it was trained to perform due to overtraining, achieving a low error rate when training, but a high error rate when testing. In both of these cases, the model cannot find a good correlation between the input and output examples.

However it is not complicated to avoid underfitting of a model. A trivial way underfitting can be avoided is to increase the training samples. In the case this is not possible due to the small size of the dataset, techniques such as data augmentation (Section~\ref{sec:unet}) in order to modify the currently existing data can be employed. Overfitting can also be avoided by providing more data, but with a higher diversity, or by performing dropout. That is, to explicitly disable perceptrons within the model in order to prevent a network to become more biased on training samples.


\subsection{Common layers of CNN}\label{sec:ml-op}

This section describes some of the most common layers that are found in Artificial Neural Networks, which is absolutely mandatory before we present the different topologies of Convolutional Neural Networks. Usually, such operations are already implemented and ready-to-use in modern Machine Learning frameworks such as TensorFlow~\cite{tensorflow2015-whitepaper}, PyTorch~\cite{paszke2019pytorch}, Caffe~\cite{jia2014caffe} or Flux~\cite{innes2018flux}.

\subsubsection{Pooling}\label{sec:pooling}

\begin{figure}[h!]
  \begin{center}
    \includegraphics[scale=1.7]{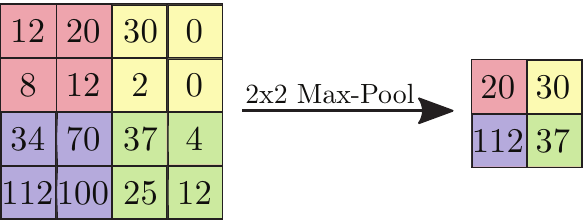}
    \caption{Max-pooling with a window size of 2x2 and a stride of 2.}\label{fig:max-pooling}
  \end{center}
\end{figure}

Pooling is an operation that achieves subsampling (i.e.~the reduction of data, by extracting only some parts of it) using a sliding window of a pre-defined size. Three popular variants of pooling are: max-pooling, min-pooling and average-pooling. Figure~\ref{fig:max-pooling} illustrates the max-pooling operation: the sliding window is moving over the input data (a matrix structure in the case of Convolutional Neural Networks), extracting the maximum value from a sub-matrix with the width and height that corresponds to the window of the pooling operation. Since we are talking about a sliding window implementation, the number of positions a window moves over the input matrix is also called the ``stride'' size. Typically, this stride size is usually equal to the width or height of the window so that there are none overlaps or gaps.

This example should suffice for the pooling operation. It should be evident from the naming that the operations of min or average pooling extract the minimum, and respectively, the average of the sub-matrix, sharing the same working principle.

\subsubsection{Convolution}\label{sec:conv}

\begin{figure}[h!]
  \begin{center}
    \includegraphics[scale=0.80]{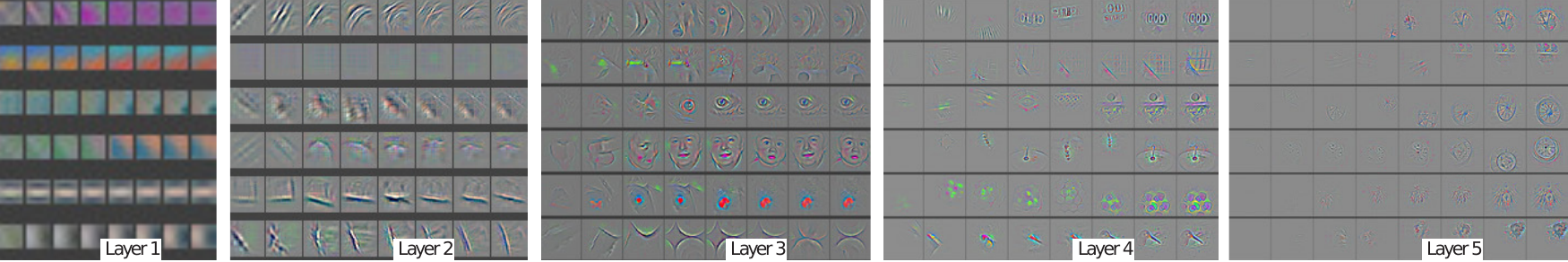}
    \caption{Feature maps learned after convolution layers~\cite{zeiler2014visualizing}}\label{fig:feature-maps}
  \end{center}
\end{figure}

The layer which give \gls{cnn} their name, the convolution operation is the one in charge of learning what in the literature are called ``feature maps'' or ``activation maps''. A visualisation of such feature maps can be seen in Figure~\ref{fig:feature-maps}. The features that are detected early on within the topology of a convolutional neural network are called ``lower-level features'' and are seen as simple lines, edges. Deeper in the network, feature maps become more complex, building on lower-level features. Those end up as the features that influence the decision proces of a model.

\subsubsection{Upsampling}\label{sec:upsampling}

\begin{figure}[h!]
  \begin{center}
    \includegraphics[scale=1.2]{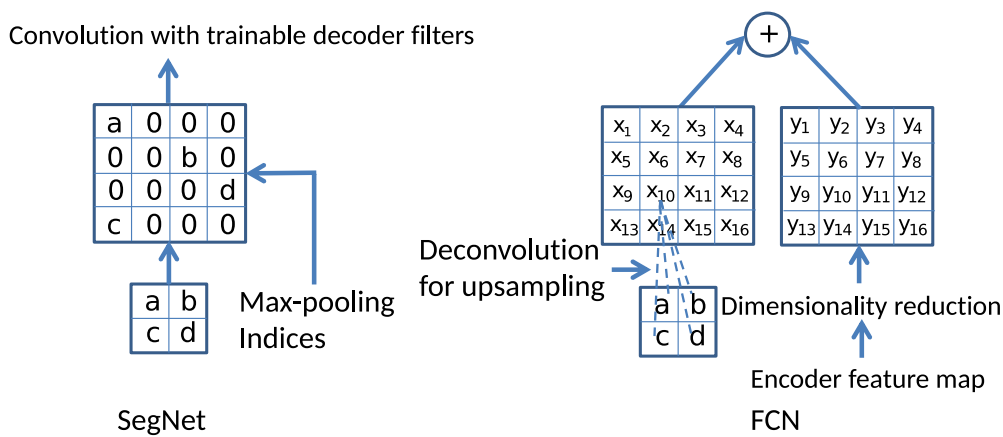}
    \caption{Max-pooling indices (left)~\cite{badrinarayanan2015segnet} | deconvolution(right)}\label{fig:maxpooling-indices}
  \end{center}
\end{figure}

The purpose of upsampling is to resample the data back to it's initial sampling rate. The CNN architectures described in this chapter present two different methods of achieving upsampling. This operation is sometimes called ``zero stuffing'' due to the fact that missing values are replaced with zero-values\footnote{This can be also seen as interpolation.} in order to perform the upsampling. Two main ways this process is performed is either through the reverse of convolution (deconvolution) or like in the case of SegNet, through reusing the indices from max pooling. Figure~\ref{fig:maxpooling-indices} illustrates the usage for both these techniques.

\subsubsection{Softmax}\label{sec:softmax}

Softmax (or softargmax) is a layer which typically found at the end of an~\gls{cnn}. It normalises and converts a vector containing real numbers into a probability distribution with the number of probabilities equal to the number of elements in the vector. Due to the nature of softmax being the last layer in a network, it is usually also called the softmax classifier. Softmax takes the form of:

$$ s(x_i) = \dfrac{e^{x_i}}{\sum_{j=1}^n e^{x_j}} $$

\subsubsection{Activation Functions}\label{sec:activation-funcs}

Part of perceptrons (Figure~\ref{fig:perceptron}) found in a typical \gls{ann}, activation functions help in deciding if a certain perceptron will take part further into the network (that is, it will propagate it's output to the next layer, hence called an ``activated perceptron''). Activation functions can be either linear or nonlinear. It is not possible to use linear activation functions with \gls{cnn}\footnote{Linear activation functions can be used to solve linear regression tasks for example.} due to the training algorithm's need to compute the derivate of such a linear function, therefore correlation to any input value will be lost, more details in Section~\ref{sec:backpropagation}. The current section showcases some of the widely used activation functions that are nonlinear and relevant to the present work.


\paragraph*{Sigmoid}

\mbox{}\\



The Sigmoid ($\sigma$) function is a widely used activation function due to it's properties: it is differentiable, monotonic, it has a smooth gradient and the output values are bound to the interval [0, 1]. A major disadvantage of Sigmoid is that at the ends of the function it's values are not very influenced by the input value, hence their derivative will converge to 0, otherwise known as the ``vanishing gradient'' problem.

\begin{center}
  \begin{minipage}{0.4\textwidth}
    \[
      \sigma(x) = \dfrac{1}{1 + e^{-x}}
    \]
    \vspace{0.5cm}
    \[
      \sigma'(x) = \dfrac{e^{-x}}{(1 + e^{-x})^2}
    \]
  \end{minipage}
  \begin{minipage}{0.5\textwidth}\raggedleft
    \includegraphics[width=\linewidth]{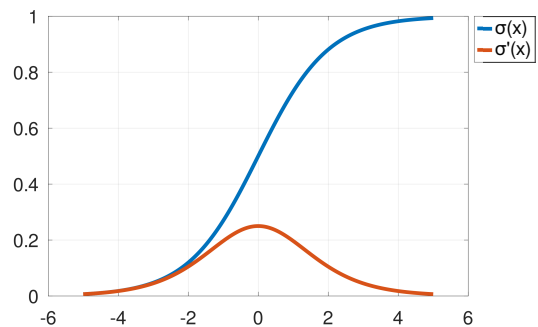}
    \captionof{figure}{The Sigmoid function}
    \label{fig:sigmoid-function-plot}
  \end{minipage}
\end{center}

\paragraph*{Tanh}

\mbox{}\\

Tanh, or the hyperbolic tangent function is similar to the Sigmoid function. Output values are bound to the interval [-1, 1] and the advantage over Sigmoid is that it has a steeper derivative, therefore the vanishing gradient problem does not appear.

\begin{center}
  \begin{minipage}{0.4\textwidth}
    \[
      tanh(x) = \dfrac{(e^{x} - e^{-x})}{(e^{x} + e^{-x})}
    \]
    \vspace{0.5cm}
    \[
      tanh'(x) = 1 - tanh(x)^2
    \]
  \end{minipage}
  \begin{minipage}{0.5\textwidth}\raggedleft
    \includegraphics[width=\linewidth]{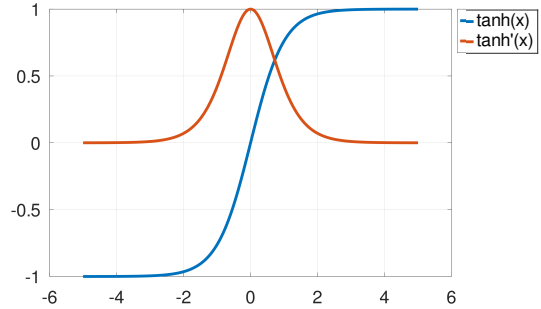}
    \captionof{figure}{The Tanh function}
    \label{fig:tanh-function-plot}
  \end{minipage}
\end{center}

\paragraph*{ELU, ReLU and Leaky ReLU}

\mbox{} \\

Exponential Linear Unit (ELU) is an activation function that also solves the vanishing gradient problem. It is highly cost-efficient therefore it leads to improving the training time of a \gls{cnn}. ELU also contains an $\alpha$ value which typically holds a small value (popular choices are between 0.1, 0.3).


\begin{center}
  \begin{minipage}{0.46\textwidth}
    \[
      ELU(x) = \left\{
        \begin{array}{ll}
          \alpha(e^x - 1),\ x\ < 0, \\
          x,\ x\ \geq 0,
        \end{array}
      \right.
    \]
    \vspace{0.5cm}
    \[
      ELU'(x) = \left\{
        \begin{array}{ll}
          ELU(x) + \alpha,\ x\ \leq 0, \\
          1,\ x\ > 0,
        \end{array}
      \right. 
    \]
  \end{minipage}
  \begin{minipage}{0.5\textwidth}\raggedleft
    \includegraphics[width=\linewidth]{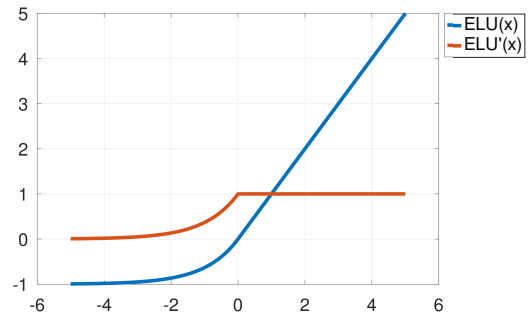}
    \captionof{figure}{The ELU function}
    \label{fig:relu-function-plot}
  \end{minipage}
\end{center}


ReLU solves the vanishing gradient problem as well, but introduces a problem called the ``dead ReLU'' - the derivative of the ReLU function provides an explanation. It's output is binary, we get either 0 or 1 values. But for values that are smaller than 0 the output will always be 0. And hence an significant number of perceptrons will not be activated, causing dropout (Section~\ref{sec:backpropagation}) in the network.

\begin{center}
  \begin{minipage}{0.4\textwidth}
    \[
      ReLU(x) = \max(0, x)
    \]
    \vspace{0.5cm}
    \[
      ReLU'(x) = \left\{
        \begin{array}{ll}
          0,\ x\ \leq 0, \\
          1,\ x\ > 0,
        \end{array}
      \right. 
    \]
  \end{minipage}
  \begin{minipage}{0.5\textwidth}\raggedleft
    \includegraphics[width=\linewidth]{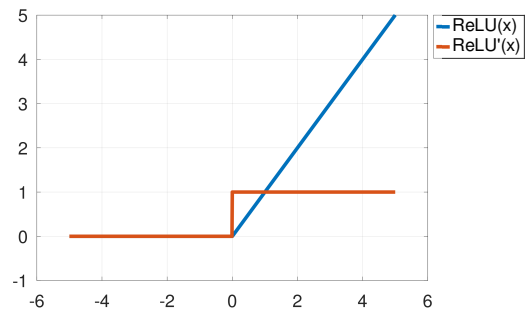}
    \captionof{figure}{The ReLU function}
    \label{fig:relu-function-plot}
  \end{minipage}
\end{center}

\pagebreak

The ``dead ReLU'' problem is fixed by Leaky ReLU. Because of the $\alpha$ value, output of Leaky ReLU cannot be 0, and is instead a small positive value in cases when the input value $x$ is less than 0. It is also faster to compute than ELU due to not having to compute the exponential function.

\begin{center}
  \begin{minipage}{0.4\textwidth}
    \[
      LReLU(x) = \left\{
        \begin{array}{ll}
          \alpha x,\ x\ \leq 0, \\
          x,\ x\ > 0,
        \end{array}
      \right. 
    \]
    \vspace{0.5cm}
    \[
      LReLU'(x) = \left\{
        \begin{array}{ll}
          \alpha,\ x\ \leq 0, \\
          1,\ x\ > 0,
        \end{array}
      \right. 
    \]
  \end{minipage}
  \begin{minipage}{0.5\textwidth}\raggedleft
    \includegraphics[width=\linewidth]{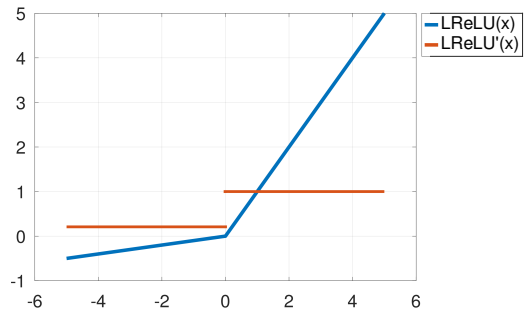}
    \captionof{figure}{The LReLU function}
    \label{fig:leaky-relu-function-plot}
  \end{minipage}
\end{center}

\subsection{Optimizers}\label{sec:optimizers}

Altough backpropagation (Section~\ref{sec:backpropagation}) has already been briefly discussed, and mentions on how the algorithm helps with computing gradients, therefore providing with information on how to adapt weights in an \gls{ann}, an optimizer is responsable with actually changing various parameters within a network. Besides changing the weights the role of an optimizer is to also changes other parameters, such as the learning rate. The learning rate is a parameter that specifies the size of a step when searching for a local minimum of a loss function (this loss function is the same one described in Section~\ref{sec:backpropagation}).

\subsubsection{Gradient Descent}

Popular with solving tasks such as linear regression or classification, and with the fact that backpropagation also makes use of gradient descent. It is an optimization algorithm which computes how weights in the network should be changed in such a way as to minimize the value of a loss function so that it can reach a global minima.

Gradient descent has a few disadvantages: namely, that it could get stuck in a local minima and not reach to find a global one. Also, the optimizer modifies the weights only after computing gradients on the whole dataset, because of this convergence to the global minima can be inefficient in terms of both time and memory usage.

\paragraph*{Stochastic Gradient Descent (SGD)}

\mbox{} \\

Stochastic Gradient Descent (SGD) is an improvement over simple gradient descent in that, parameters are updated at regular intervals, therefore requiring less time to converge to a global minima. The disadvantage is that SGD might continue to converge even after reaching a global minima due to constantly updating parameters.

\subsubsection{Adam}\label{sec:adam}

Adam (or Adaptive Moment Estimation) is an optimizer that works with momentums (which are an heuristic for improving convergence in order to find a global minima). The advantage of Adam, since momentums are used is that the convergence is typically fast, but at a high computational cost.


\subsection{Fully Convolutional Neural Networks}\label{sec:fcn}

\begin{figure}[h!]
  \begin{center}
    \includegraphics[scale=1.5]{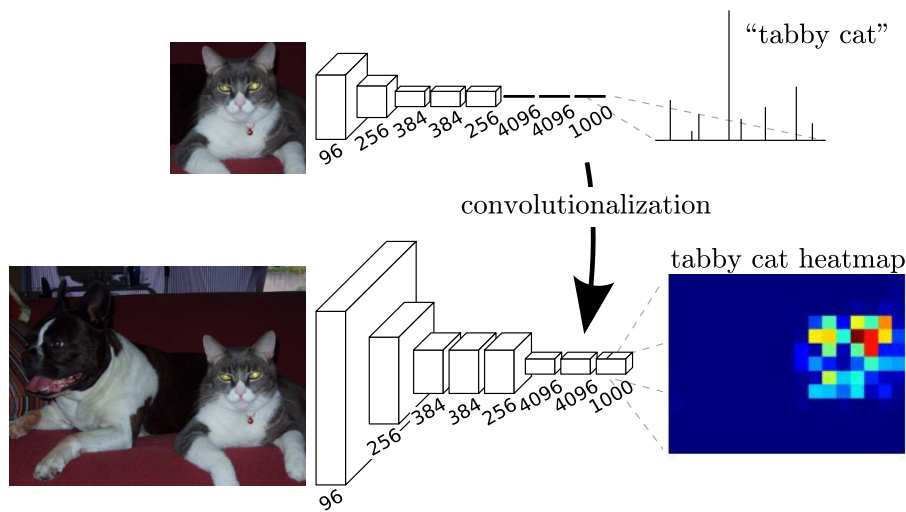}
    \caption{Fully convolutional network convolutionalization~\cite{long2015fully}}
    \label{fig:fcn}
  \end{center}
\end{figure}

Fully Convolutional Neural Networks~\cite{long2015fully} (\gls{fcn}) are different from typical~\gls{cnn} by not containing fully connected layers. The fully connecting layers are instead replaced by convolutions (Section~\ref{sec:conv}) with a filter size equal to the size of input data. This conversion allows for networks which are trained for classifying images into outputing heatmaps (such is the example seen in Figure~\ref{fig:fcn}) of the features which determine the model into attributing a certain class to the input image.

In comparison to state of the art ~\gls{cnn} topologies, the proposal of converting typical fully connected networks (this process is also called convolutionalization) ~\gls{fcn} that were designed by~\cite{long2015fully} show an increase of accuracy performance of up to 20\% compared to the same topologies which contain regular fully connected layers.

\pagebreak

\subsection{SegNet}~\label{sec:segnet}

SegNet~\cite{badrinarayanan2015segnet} is a fully-convolutional (Section \ref{sec:fcn}) Neural Network that was designed specifically with semantic segmentation in mind. The topology of this network is comprised of two parts: an encoder and a decoder. Based on VGG16~\cite{simonyan2014very}, the first 13 convolutional layers (which comprise the encoder part) of SegNet are identical: multiple convolution layers followed by batch normalization and ReLU activation and max-pooling with a 2x2 window and a stride of 2. Each layer in the encoder has a corresponding layer within the decoder where upsampling is performed.

According to~\cite{badrinarayanan2015segnet}, one of the incentives for developing SegNet is the ability to map low-resolution feature maps generated in the encoder part of the network with higher-resolution features (whch are created with the help of upsampling) into the decoder of the network, therefore making the task of boundary localization easier. This approach is one that is proven beneficial for pixelwise semantic segmentation.

\begin{sidewaysfigure}[p]
    \includegraphics[scale=0.25]{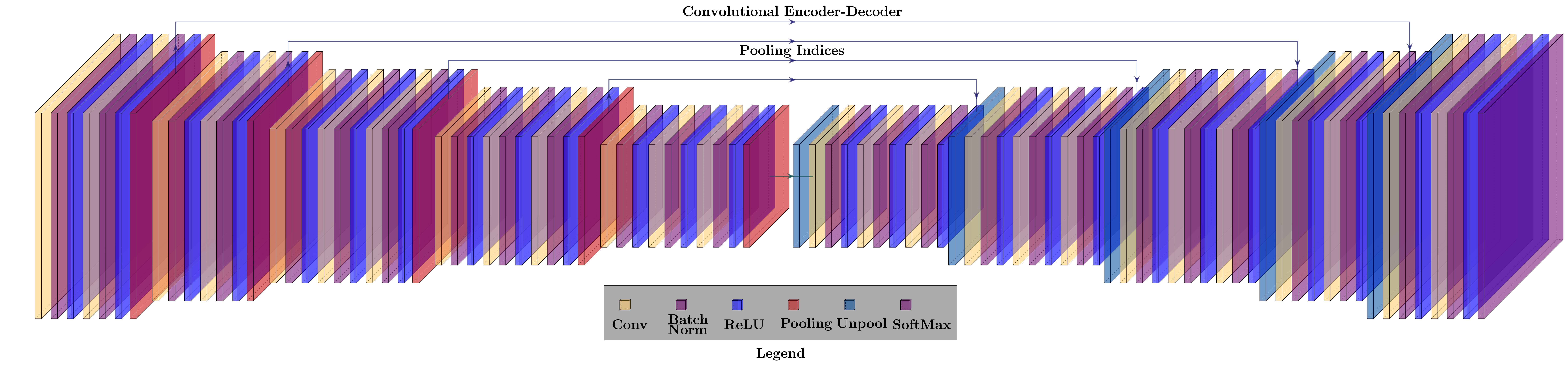}
    \caption{The SegNet architecture~\cite{badrinarayanan2015segnet}}
    \label{fig:segnet-architecture}
\end{sidewaysfigure}

Each block in the encoder part of the network contains a convolution followed by batch normalization and a ReLU activation function later followed by max-pooling for downsampling data~\ref{fig:segnet-architecture}. Fully connected layers are removed in order to free memory to contain higher resolution feature maps. Additionally, using max-pooling indices instead of keeping the entire feature maps from encoding reduces memory when inferencing data compared to other approaches of~\gls{fcn}~\cite{badrinarayanan2015segnet}.

The architecture of SegNet is differentiated from other~\gls{cnn} based on how it handles the upsampling process. Namely, max-pooling indices are reused in order to resample the image to bigger dimensions (Section~\ref{sec:upsampling}). It therefore eliminates the need of using deconvolutions (which how most networks learn to resample data).

\newpage

\subsection{U-Net}~\label{sec:unet}

U-Net~\cite{ronneberger2015u} is another \gls{fcn} model, which was created for semantic segmentation of biomedical images, specifically electron microscopy imagery. The lack of available datasets within this specific field has lead to developing a ~\gls{cnn} which is able to perform well on small datasets, while heavily relying upon augmentation techniques.

\begin{sidewaysfigure}[p]
  \begin{center}
    \includegraphics[scale=0.55]{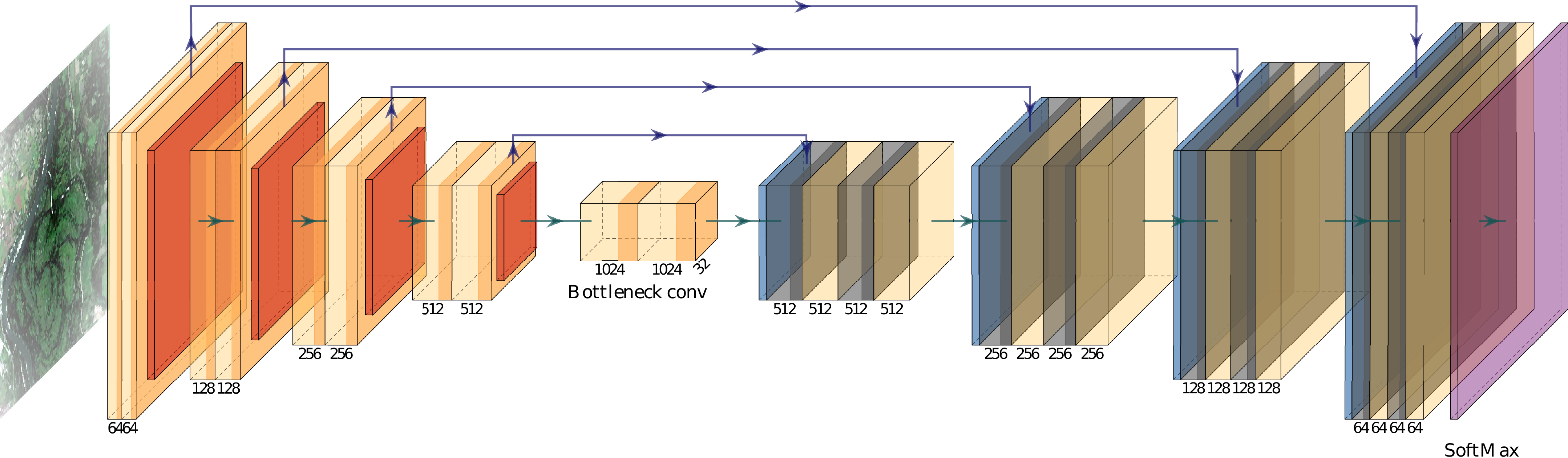}
    \caption{The U-Net architecture~\cite{ronneberger2015u}}
    \label{fig:unet-architecture}
  \end{center}
\end{sidewaysfigure}

U-Net also uses an encoder-decoder topology, and is comprised of 32 convolution layers. The layers found in the U-Net architecture are similar to the ones in SegNet: 3x3 convolutions, followed by ReLU activation and max-pooling with a 2x2 window and stride size of 2. Every downsampling part in the encoder doubles the number of features found in the network. U-Net performs upsampling by deconvolution (Section~\ref{sec:upsampling}) that reduces the number of feature channels in half. Feature maps from the corresponding layer in the encoder are cropped and concatenated to the ones found in the decoder, followed by 3x3 convolutions and ReLU. The network uses a 1x1 convolution at the end for mapping each output vector to a class~\ref{fig:unet-architecture}. This 1x1 convolution can be usually replaced by the softmax function~\ref{sec:softmax}. The motivation for cropping and concatenating the feature maps is that of propagating information faster in the network.

U-Net was trained for solving semantic segmentation on electron microscopy images. Since the dataset was relatively small in samples, data augmentation techniques were used, such as: deformations, rotation and shifts. Additionally, U-Net has already been used with Remote Sensing data~\cite{buslaev2018fully,he2016deep,van2018spacenet} yielding satisfactory results.

\pagebreak

\subsection{ResNet}~\label{sec:resnet}

ResNet~\cite{he2016deep} proposes a solution for creating models that are resistant to the problem of accuracy degrading over time. This problem usually occurs when using deep neural network topologies and has the effect of restricting the number of layers of a network as to not produce high error rates. The proposed method is the introduction of residual units (Figure~\ref{fig:residual-unit}) in the network. This approach has been verified by adapting deep neural networks and training on datasets such as ImageNet~\cite{deng2009imagenet}, CIFAR-10~\cite{krizhevsky2009learning} or MS-COCO~\cite{lin2014microsoft}. Improvements over training time and overall model accuracy while using the aforementioned datasets have also been shown~\cite{he2016deep}.

\subsubsection{Residual Units}

A residual unit (also known as a ``skip connection'') is a shortcut between layers of an \gls{ann}. Adding skip connections helps with avoiding the vanishing gradient problem: when training the gradient produced by backpropagation for a given weight can becomes small, hence making weights stop updating their values. The proposal with ResNet~\cite{he2016deep} helps backpropagation propagate gradients through the skip connections more conveniently. Using residual units with U-Net and other topologies has been shown to reduce the number of parameters and training time~\cite{zhang2018road}.

\begin{figure}[h!]
  \begin{center}
    \includegraphics[scale=0.35]{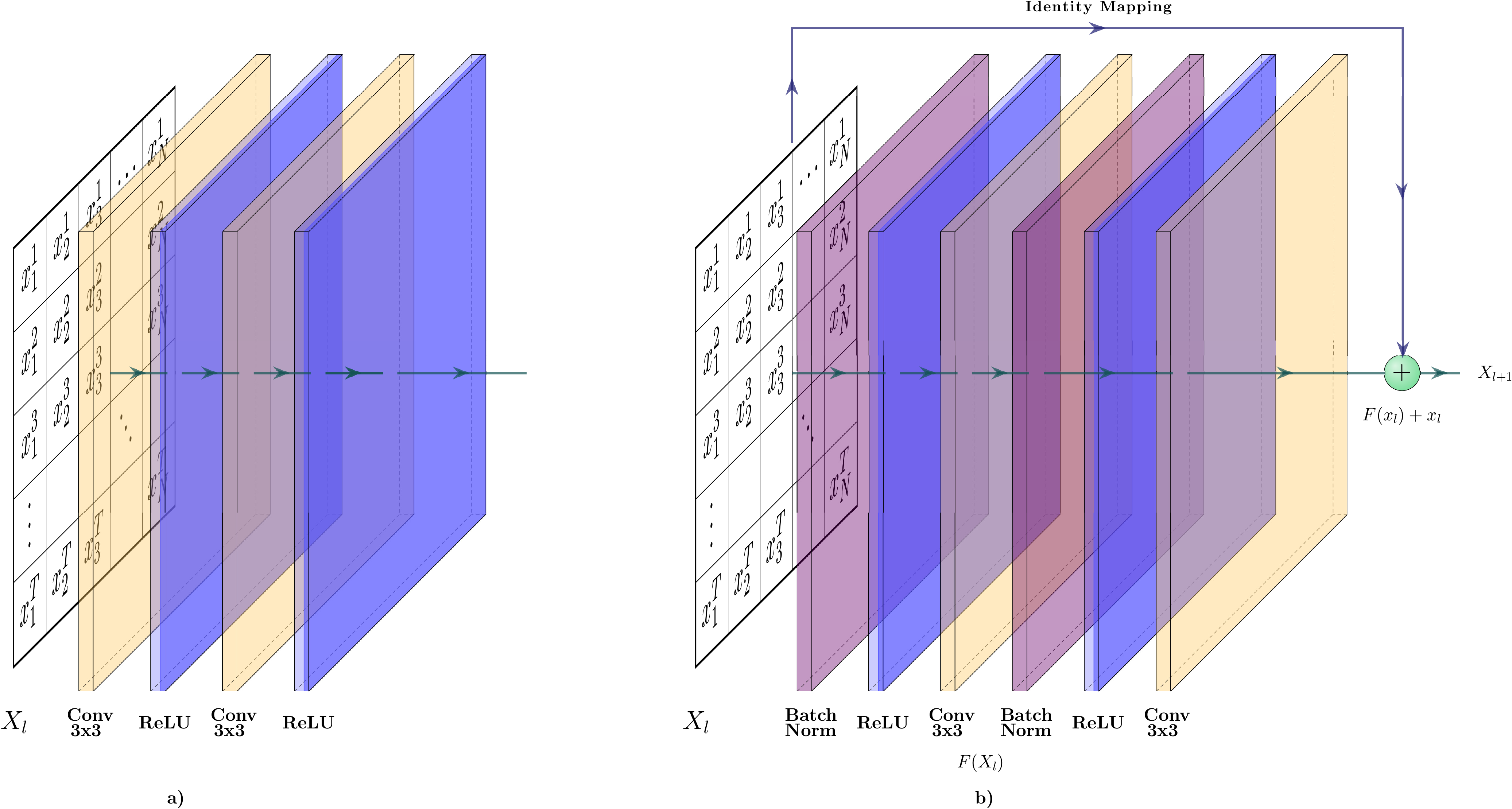}
    \caption{a) Simple convolution layers | b) Residual Unit~\cite{he2016deep}}
    \label{fig:residual-unit}
  \end{center}
\end{figure}

Concatenate the output from previous layers is performed by using an identity mapping (Figure ~\ref{fig:residual-unit}). In the case of an difference between dimensions of a layer that needs to be concatenated (due to having convolutions or pooling layers) either zero-stuffing or the use of 1x1 convolutions is employed. Figure~\ref{fig:residual-unit} shows an residual unit compared to a typical convolution block found within an typical~\gls{cnn}.

\section{Measuring the performance of models}~\label{sec:performance}

There are numerous means of quantifying the performance yielded by a Machine Learning algorithm. This section is dedicated to showcasing some of those methods. Beforehand, we should introduce the notions of True Positives, True Negatives, False Positives and False Negatives. This is shown within Table~\ref{tab:confusion-matrix}, also called a ``confusion matrix''. Varying from what a Machine Learning model predicts and what the actual class is\footnote{In Remote Sensing, the actual class is also called the ``ground truth''.}, the result is classified into one of these four categories.

\vspace{0.5cm}

\begin{tabular}{l|cc}\label{tab:confusion-matrix}
  & \textbf{Predicted Positive} & \textbf{Predicted Negative} \\
  \hline
  \textbf{Actual Positive} & \textbf{T}rue \textbf{P}ositive & \textbf{F}alse \textbf{N}egative \\
  \textbf{Actual Negative} & \textbf{F}alse \textbf{P}ositive & \textbf{T}rue \textbf{N}egative \\
\end{tabular}

\subsection{Accuracy: is it enough?}

Accuracy is a very commonly used way of measuring how many samples a Machine Learning model can correctly predict out of the whole validation dataset.


$$ accuracy = \dfrac{TN + TP}{TP + FP + TN + FN} $$

While accuracy can give us an insight of the number of samples a given model correctly predicts, there exists one problem: It might not be representative to the problem. This is the case when True Positives highly outnumber True Negatives or vice versa. In scenarios where, for example, a model would learn to perform the task classification, relying on accuracy only as an evaluation metric might be problematic, we could have a high accuracy rate (such as $\geq$ 90\%), but what our model might be doing is actually classifying all the samples as the class corresponding to True Negatives.~\textit{In other words, if we have 10.000 True Positives, and 1.000 True Negatives samples, a model that would classify all the samples as True Positives, will have an accuracy of 91\%, which does not sound bad at all, but our model will only be classifying one class}. Taking all of this into consideration, accuracy cannot be always relied upon as the only means of evaluating the performance of a model.

\subsection{Precision \& Recall}\label{sec:precision-recall}

Given the problem discussed in the previous section regarding accuracy, there exist other metrics which can be used as an alternative to keep track of the performance yielded by a~\gls{ml} model, namely precision and recall. Precision is represented as:

$$ precision = \dfrac{TP}{TP + FP} $$

From the above formula, we can understand that precision is used as a means of identifying how many of the samples our model classified as positive (TP + FP), are indeed actual positives samples.~\textit{If a model correctly classifies 910 samples (either true or false positives) out of 1000 true positives, it's precision would be 91\%}.

The recall metric, on the other hand tells us how many samples that are relevant (TP + FN) are classified appropriately by our model.~\textit{If a model classifies 940 samples (either true positives or false negatives) out of 1000 true positives or false negatives, it's recall is 94\%}.

$$ recall = \dfrac{TP}{TP + FN} $$

\subsection{F-score / Sørensen-Dice coefficient}

This metric is taking into consideration both precision and recall for measuring the accuracy of a ~\gls{ml} model. The key difference between this metric and the simple accuracy shown previously, is that F-score is the weight added to both false negatives and false positives. Hence, larger numbers of true negatives will not have as big of an impact as in the case of simple accuracy. F-1 score is expressed as:




$$ F_1 = 2 * \dfrac{precision * recall}{precision + recall} $$


The Sørensen-Dice coefficient (DSC) is a particular case of the $F_1$ score, when either one of precision or recall has a value of 0, and it can be written as:

$$ DSC = \dfrac{2TP}{2TP + FP + FN} $$

\subsection{IoU/Jaccard index \& MIoU}\label{sec:miou}


MIoU or the Jaccard index is a metric used typically in measuring the performance of models for solving Computer Vision tasks.

$$ Jaccard(X,Y) = \dfrac{|X \cap Y|}{|X \cup Y|} $$

Where the X and Y sets represent either bounding boxes, or segmentation masks for the ground truth (X), and predictions performed by the model (Y).




\pagebreak

\section{Multi-modal learning}\label{sec:multi-modal}

Multi-modal learning~\cite{baltruvsaitis2018multimodal} employs the principle that multiple different kinds of data from multiple sources are used in order to develop an automated model that is capable of performing on a given task. Similar to what a human experiences, that is recieving multiple information from different sources (through systems such as vision, auditory, olfactory, sensory). In the context of~\gls{ml} the nature of the data dictates how learning representations should be achieved. 

One of the main challanges of multi-modal learning is that of correlating data (also known as alignment). In the case of Remote Sensing, alignment has to be done both temporal and spatial to assure relevance of the data from an holistic standpoint. Other multi-modal learning techniques encompass the correlation of text audio and/or images. When dealing with multiple sources that produce similar data (such as imagery that is represented as a matrix structure) there are multiple posibilities of achieving multi-modal training. Additionally, learning representations (Section~\ref{sec:ai-techniques}) from data spanning multiple different sources increases the complexity of multi-modal learning. Data fusion is the ability to aggregate the information in such a way that one or multiple models are capable of yielding informed predictions.


Using multiple-input~\gls{cnn} topologies is a method of achieving multi-modal training. The working principle with multiple inputs is that data is fed to the network on separate input layers and is to be fusioned later into the topology. This process is usually performed by using typical operations of~\gls{cnn} (described in Section~\ref{sec:ml-op}).

Ensemble learning~\cite{zhou2021ensemble} (otherwise known as ``commitee-based learning'') is another method multi-modal learning can be performed. The principle is to train separate models using data from separate sources and to use a consensus algorithm after predicting in order to reach a decision, taking into consideration the features learned by each model based on the data that was given.

In the current work, three different sources of data are aggregated in order to build a dataset designed for multi-modal learning (Section~\ref{sec:my-dataset}). Data gathered by the Sentinel-3~\gls{olci} instrument is supposed to provide additional information to the Sentinel-2~\gls{msi} data. Additionally~\gls{clc} data further enhances the available information, providing ground truth which necessary for performing supervised learning.

\pagebreak

\section{Introducing Hugin: Machine Learning using remote sensing data}~\label{sec:hugin}

Hugin\footnote{\url{https://pypi.org/project/hugin/}} is a \gls{ml} framework specifically designed to work with multiple geospatial raster data formats. Currently built on top of the Keras\footnote{\url{https://keras.io/}} library, it provides support for solving~\gls{cv} tasks such as classification, semantic segmentation or super-resolution and data fusion. It allows for training~\gls{cnn} models with data acquired by Remote Sensing techniques (e.g. those discussed within Section~\ref{sec:rs-datasets}).

Hugin is highly configurable (Appendix ~\ref{sec:hugin-configs}) based on YAML\footnote{\url{https://yaml.org/}} files for explicitly controlling both the training and testing processes. Included in these configuration files are data sources (with support for local or cloud based object storage such as Amazon S3 Buckets\footnote{\url{https://aws.amazon.com/s3/}}). Hugin can correlate multiple data sources, load models from paths to custom built or existing Keras model builders, change parameters to these models such as optimizers (Section~\ref{sec:optimizers}) and their respective parameters, loss functions, performance monitoring metrics (Section~\ref{sec:performance}). Control over introducing Keras callbacks that aid the training process is also included, some examples such as as stopping training a model if a specified treshold of change in a given metric is not met after a converging for a while or heuristics for reducing the learning rate.

When dealing with datasets with particularly large images, Hugin can also perform image tiling automatically by an implemented sliding window algorithm which can be configured by using the YAML files. It also provides aid when data is too large to fit into memory when training, therefore loading data in predefined batches.

Checkpointing is a useful feature that has been added within Hugin for saving the weights of a model either at specified intervals, or saving only the weights of the model which achieves the best overall performance in terms of a monitored metric.

Predictions are also configurable with Hugin, with support for both unimodal and ensemble predictions. The output of predictions is controlled by using either a specialised raster writer such as RasterIO\footnote{\url{https://rasterio.readthedocs.io/en/latest/}} or by wrapping custom format exporters. Similar to training configuration, predictions can also be specified which model loaders, and data sources to use.


%% file: sections/approach.tex
\chapter{Semantic segmentation of vegetation in Remote Sensing imagery: an approach based on Machine Learning}\label{ch:approach}

\chaptermark{An approach based on machine learning}

\begin{figure}[h!]
  \begin{center}
    \includegraphics[scale=0.7]{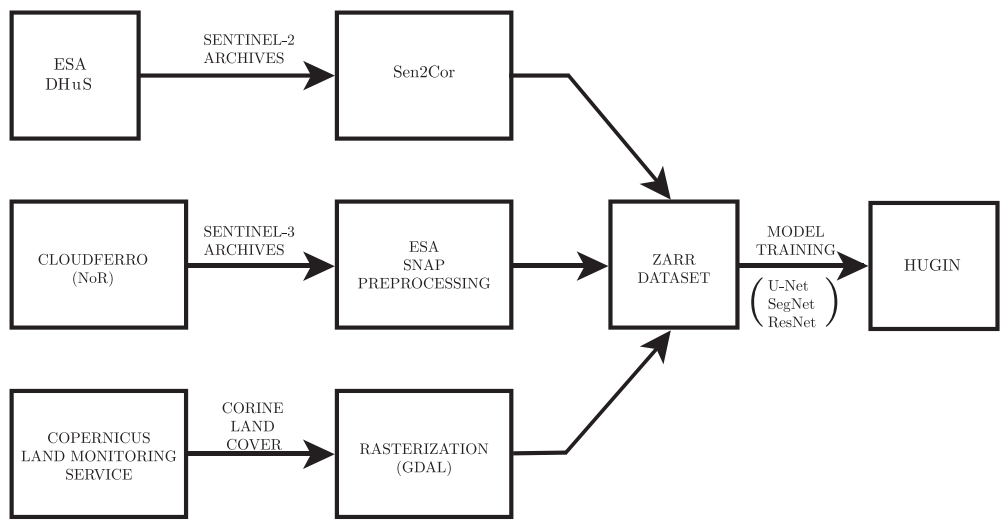}
    \caption{The processing pipeline}
    \label{fig:pipeline}
  \end{center}
\end{figure}

Figure~\ref{fig:pipeline} illustrates the pipeline that was used for solving the problem of semantic segmentation of vegetation classes in the current approach. Three different data sources are used: \gls{esa}~\gls{dhus}, the CloudFerro\footnote{\url{https://cloudferro.com/en/}} cloud provider, and the Copernicus Land Monitoring Service\footnote{\url{https://land.copernicus.eu/}}. Data from these sources was each later preprocessed in a specific way, and stored using the hierarchical zarr\footnote{\url{https://zarr.readthedocs.io/en/stable/}} data storage format. This data was later used in order to train three different configurations of ~\gls{cnn} (which are described in Sections~\ref{sec:segnet},~\ref{sec:unet},~\ref{sec:resnet}) with the help of the Hugin~\gls{ml} framework (Section~\ref{sec:hugin}). Specifics regarding the training process are described within Section~\ref{sec:hugin-training}, and the results of applying these methods are presented in Chapter~\ref{ch:experiments}.

\section{ESA DHuS as a data source}\label{sec:dhus}

Satellite imagery gathered by the Copernicus programme is distributed through the \gls{esa} Ground Segment\footnote{\href{https://www.esa.int/Applications/Observing_the_Earth/Copernicus/Ground_Segment_overview}{\nolinkurl{esa.int/Applications/Observing_the_Earth/Copernicus/Ground_Segment_overview}}}. Data is free to use and to redistribute and is available within any of the multiple Data Hub Software (\gls{dhus}). \gls{dhus} nodes for distributing data that are synchronized with official \gls{esa} instances can be operated by entrusted\footnote{By the European Commision (\gls{ec})} entities of the Copernicus programme, those instances are called collaborative nodes.

\subsection{Extracting data from DHuS}

Each \gls{dhus} instance provides with two methods for searching Sentinel products: a web interface, and an OpenSearch\footnote{\url{https://scihub.copernicus.eu/userguide/OpenSearchAPI}} \gls{api}. The \gls{api} includes means of filtering data based on sensing period, Sentinel mission, product type, processing level, percentage of cloud coverage or missing data. Support for filtering data by geographical location is also available, provided the usage of a markup language specifically designed for representing geometrical objects that are georeferenced: Well Known Text (\gls{wkt}).

Since the web interface does not support for downloading data in bulk, we want to programatically extract Sentinel archives through the available OpenSearch API (an example of usage can be found in Appendix~\ref{sec:dhus-opensearch}). The Sentinelsat\footnote{\url{https://github.com/sentinelsat/sentinelsat}} project provides an implementation of an \gls{dhus} OpenSearch client that is easily configurable for searching and retrieving archives from all the Sentinel missions found within~\gls{dhus}.

Sentinelsat was used for filtering Sentinel-2 archives that we're acquired on the territory of Romania for the year of 2018 in order to keep relevance with the land cover classes in the ~\gls{clc} inventory. The filtered archives we're then retrieved and processed using Sen2cor for extraction of cloud masks and cloud shadow masks~\ref{sec:cloudmasks}.

\subsection{Current limitations of \gls{dhus} and alternatives}

Within the~\gls{dhus} services a rolling plan was introduced\footnote{\url{https://scihub.copernicus.eu/userguide/LongTermArchive}} in which older archives are placed inside of a long-term archive (\gls{lta}). The data is still available for retrieval by users within 24 hours per archive, on a by request basis that is limited to one archive per day. Altough \gls{esa} sustains a great ammount of data, the rate at which data is being produced and the infrastructure requirements for serving archives have grown recently in such a manner that storing older archives in \gls{lta} has become mandatory.

Additionally with the growth of data, \gls{esa} has developed the ~\gls{eo}~\gls{nor} initiative, supporting research and development with the help of subcontracting partners which provide \gls{rs} data and processing services in the cloud. To work around the~\gls{lta} issue, the current approach was sponsored for fulfilling a~\gls{nor} request with the scope of retrieving and processing Sentinel-3~\gls{olci} archives in order to complete the dataset. The retrieval and processing of Sentinel-3~\gls{olci} archives for the year of 2018 has been accomplised in the CloudFerro cloud, which besides cloud computing capabilities it also provides a collection of unarchived Copernicus mission imagery.

\vfill

\pagebreak

\section{Sentinel-2 and Sentinel-3 archives}

\subsection*{Sentinel-2}\label{sec:sentinel2-description}

Currently with two satellites in orbit, Sentinel-2A and Sentinel-2B which together have a revisit time of 5 days (meaning they pass through the same geographic location to acquire data at an interval of 5 days). The Sentinel-2 products provide~\gls{msi} data at three diferent spatial resolutions (60m, 30m, and 10m) covering a wide spectrum.~\gls{msi} reflectance data is important for providing spectral signatures of different objects which can be used for classyfing different objects. The contents of a Sentinel-2~\gls{msi} archive are listed in Appendix~\ref{sec:s2-data-msi}. Sentinel-2 products for 2017 from the~\gls{aoi} are included in the proposed dataset (described fully in Section~\ref{sec:my-dataset}).

\subsection*{Sentinel-3 OLCI}\label{sec:sentinel3-olci}

The~\gls{olci} instrument on board of the two Sentinel-3 (Sentinel-3A and Sentinel-3B) satellites provides with additional information about land coverage at a revisit time of two days. This information is useful for enhancing Sentinel-2~\gls{msi} data and for performing multi-modal learning (Section~\ref{sec:multi-modal}). Contents of the Sentinel-3~\gls{olci} archives (which are listed within Appendix~\ref{sec:s3-data-olci}) include as previously mentioned (in Section~\ref{sec:rs-instruments}), reflectance data for all of the 21 bands of the~\gls{olci} instrument, tie geometries, high resolution georeferencing, meteorology data, removed pixels informations and quality control flags.

\section{Data preprocessing}

Before constructing the dataset and performing multi-modal learning, data from all the aforementioned sources has to be preprocessed. In the context of the current work, this includes reprojection of Sentinel-3 archives so we can correlate them to Sentinel-2 and~\gls{clc} data. Since~\gls{clc} is distributed in vectorial format, it needs to be rasterized at the same spatial resolution as the Sentinel-2 data in order for alignment of input and output data to be possible.

\subsection{Reprojecting Sentinel-3 archives using ESA GPT}\label{sec:gpt-processing-cloudferro}

Sentinel-3~\gls{olci} archives include reflectance data for each band in separate files. The georeferencing is also stored in a separate file. For relevance, we want to include georeferencing information and store all the bands in a single master file. Additionally, bands information in a Sentinel-3 archive are provided in a NetCDF-4\footnote{\url{https://www.loc.gov/preservation/digital/formats/fdd/fdd000332.shtml}} format, for uniformity and similarity to the other data, we want to rewrite the data into GeoTIFF\footnote{\url{https://www.loc.gov/preservation/digital/formats/fdd/fdd000279.shtml}} format. The~\gls{esa}~\gls{snap} (Sentinel Application Platform) provides multiple toolkits for interacting with data acquired by the Copernicus programme. Processing graphs (example in Appendix~\ref{sec:gpt-processing}) can be described using XML format and executed with the help of the Graph Processing Tool (\gls{gpt}).

\subsection{Corine Land Cover rasterization}\label{sec:rasterization}

Corine Land Cover~\gls{clc} is distributed in vector format (there exist a raster which is distributed as well, but it was rasterized at 100 meter resolution). To achieve a finer granularity, the vector \gls{clc} data which is distributed under the proprietary ESRI OpenFileGDB format is rasterized using the GDAL\footnote{\url{https://gdal.org/}} \textit{gdal\_rasterize} tool at a resolution of 10m. The output of this process is a single GeoTIFF file containing all the information from initial the~\gls{clc} vector. Since our~\gls{aoi} is the entire territory of Romania, the output raster was cropped to a bounding box which is expressed as an~\gls{wkt} polygon and passed to \textit{gdal\_rasterize} that delimits the whole~\gls{aoi}.

\subsection{Cloud masks - Sen2Cor}\label{sec:cloudmasks}

Sen2cor~\cite{main2017sen2cor} is a processor for Sentinel-2 Level-2A~\gls{boa} reflectance products. The processor is present within the~\gls{snap} toolbox for Sentinel-2, and it can perform classification (SCL) of 11 classes (with two important classes which are belonging to cloud and cloud shadows - medium and high cloud probabilities). It can be used to also perform atmospheric corrections in Sentinel-2 archives. Sen2cor was used in the context of the current work for generating cloud masks, in order for regions of imagery that are occluded (Section~\ref{sec:cv}) by clouds to be ignored by the model and not taken in as samples of a class present in the~\gls{clc} inventory.


\subsection{Tiling and mosaicking}

In order to obtain a finer granularity (in terms of selecting data from a specific geographic location, but also being memory efficient) making data easier to handle, in the case of~\gls{rs} data it is common to divide rasters into smaller chunks called tiles. This was the approach taken with the proposed dataset. Sentinel-2~\gls{msi} data was chunked into tile sizes of 256x256 with 4 dimensions (one for each band) for the bands with 10m resolution, 6 dimensions for bands at 20m and 3 for 60m. Sentinel-3~\gls{olci} data was chunked in 9x9 tiles with 21 dimensions, since all the bands have the same spatial resolution of 300m.

For visualisation of scenes or for retrieving or computing global information in a particular area of interest, scenes are assembled based on their georeferencing into a mosaic. Several technologies including virtual rasters supported by \gls{gdal} can compose an aggregate of multiple rasters.

\section{Constructing a multi-modal dataset}\label{sec:my-dataset}

All the processing steps of the pipeline illustrated in Figure~\ref{fig:pipeline} for the data spanning three different sources have been described. Having as preprocessed output rasters for both Sentinel-2~\gls{msi} and Sentinel-3~\gls{olci} stored in GeoTIFF format, data is tiled and added into a hierarchical data storage format, namely the zarr format.

\subsection{Zarr as a data storage format}

Zarr is a hierarchical data format which is similar to \gls{hdf}\footnote{\url{https://portal.hdfgroup.org/display/HDF5/HDF5}}. It has support for storing chunked and compressed (using the blosc\footnote{\url{https://github.com/Blosc/c-blosc}} algorithm by default, but data can also be compressed with the zlib, bz2 and lzma algorithms) multi-dimensional arrays spanning multiple different numpy\footnote{\url{https://numpy.org/}} data types. Since zarr is a hierarchical data storage format, it has support for creation of groups and arrays within groups.

\begin{figure}[h]
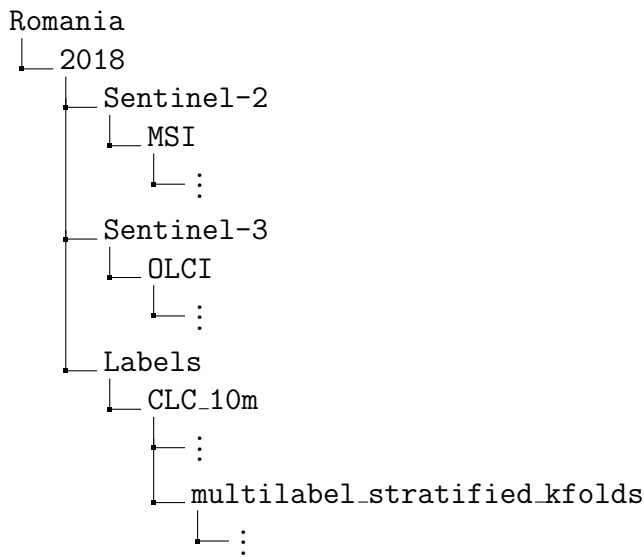

  \begin{minipage}{0.9\textwidth}
    \dirtree{%
      .1 Romania.
      .2 2018.
      .3 Sentinel-2.
      .4 MSI.
      .5 \vdots.
      .3 Sentinel-3.
      .4 OLCI.
      .5 \vdots.
      .3 Labels.
      .4 CLC\_10m.
      .5 \vdots.
      .5 multilabel\_stratified\_kfolds.
      .6 \vdots.
    }
  \end{minipage}
\caption{Directory tree for the zarr structure of the proposed dataset.}
\label{fig:zarr-dataset}
\end{figure}

The proposed structure for storing the Sentinel archives and the~\gls{clc} inventory presented within the current work under zarr format is illustrated in Figure~\ref{fig:zarr-dataset}. It was designed in such a way to enable further enhancements and addition of data.

\vspace{0.3cm}

\begin{figure}[h!]
  \begin{center}
    \includegraphics[scale=0.45]{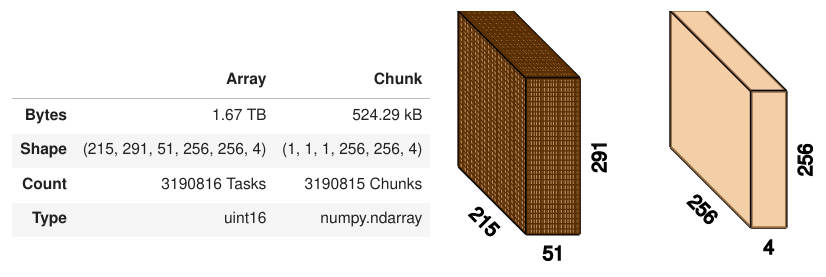}
    \caption{Sentinel-2 10m resolution data array for 51 weeks of year 2018.}
    \label{fig:s2-dataset-2018}
  \end{center}
\end{figure}

Figure~\ref{fig:s2-dataset-2018} illustrates the array containing Sentinel-2~\gls{msi} 10m bands data for 51 weeks of the year 2018. It has a total size of 1.67TB, spanning on a width of 215 tiles and a height of 291 tiles (this width and height cover the whole~\gls{aoi}). Each tile has a size of 256x256 pixels and 4 channels (Red, Green, Blue and Near Infrared).

\subsection{Splitting the dataset: Stratified K-folds}\label{sec:kfolds}

Simple methods for splitting the dataset into training and testing subsets, such as the \textit{sklearn.model\_selection.train\_test\_split}\footnote{\href{https://scikit-learn.org/stable/modules/generated/sklearn.model_selection.train_test_split.html}{\nolinkurl{https://scikit-learn.org}}} implemented in the scikit-learn library perform shuffling and splitting of the dataset in two sets, based on user defined proportions, a popular choice is using 80\% for training and 20\% for testing.



In order to avoid overfitting (described in Section~\ref{sec:under-overfitting}) of any models we wish to train, technique was utilized for cross validation. K-fold cross validation gives information about how a model will be able to perform on samples it has not seen.

K-fold is an algorithm for splitting the dataset into subgroups of equal size for training and validation of models, which makes better use of the samples within a dataset by keeping each subgroup statistically representative of the entire dataset. The working principle of the K-fold algorithm are shown within Algorithm~\ref{alg:kfold-algo}.

\begin{algorithm}
  \caption{K-fold algorithm}
  \label{alg:kfold-algo}
  \KwData{Input: $K > 1$\Comment*[r]{Number of groups}}
  \KwData{Input: $\Theta$\Comment*[r]{Set of hyperparameters}}
  \KwData{Input: D\Comment*[r]{Initial dataset containing input and output data}}
  \KwData{Input: M\Comment*[r]{Trainable model}}
  $score = 0$\;
  partition D into K randomly equal size subsets $F_1 \ldots F_K$\;
  \ForEach{$\theta \in \Theta$}{
    \For{i = 1 to K}{
      Use $F_i$ as a validation dataset and the remaining folds for training\;
      Fit data to model M using hyperparameter configuration $\theta$\;
      Compute training error \textbf{e} of model M with for the holdout fold $F_i$\;
    }
    $score \mathrel{+}=\dfrac{e}{K}$\;
  }
  \KwRet{score}
\end{algorithm}

As illustrated in Algorithm~\ref{alg:kfold-algo} in the implementation of the K-fold algorithm, each subset $K_i$ is used at least once for validation. This means that each sample in the dataset will be used once for validation and $K-1$ times for training the model $M$.

There exist different techniques for choosing the value of $K$: Picking a value that would make sure that each validation and training set is large enough to be statistically representative for the entire dataset. $K=10$ is a popular choice in the~\gls{ml} community. $K=n$ where $n$ is the size of the dataset (this is called ``leave-one-out cross validation'') will validate on one sample and use the rest folds for training.

Stratified cross validation with the k-fold algorithm is typically used when dealing with multiple labels (multilabel stratified k-fold). It ensures that when the dataset is split into subsets, each subset besides having the property of being equal to the other $K-1$ sets, they will contain approximately the same number of samples of each class. Using the stratified variation of k-fold therefore preserves class ratio.

\section{Training Convolutional Neural Networks with Hugin}~\label{sec:hugin-training}

The~\gls{cnn} architectures presented within Sections~\ref{sec:segnet}, \ref{sec:unet} and \ref{sec:resnet} were used in order to train models on the task of semantic segmentation using 17 vegetation related classes of the proposed multi-modal dataset (Section~\ref{sec:my-dataset}).

In terms of configurations, each model was trained with the Adam optimizer~\ref{sec:adam} using categorical cross-entropy as a loss function. The metrics used for monitoring model performance are \gls{miou} (Section~\ref{sec:miou}), precision and recall (Section~\ref{sec:precision-recall}).

Categorical cross-entropy (which is Softmax + cross-entropy) was chosen as a loss function since our goal is to perform semantic segmentation for multiple classes of vegetation. The Adam optimizer was chosen to provide faster convergence, since training time grows significantly with the number of classes involved.

Keras callbacks such as early stopping and reducing of the learning late when loss value on the validation dataset reaches a plateau were used in order to achieve better performance of the models. Checkpointing implemented in~\ref{sec:hugin} was used to save only the weights for the epochs which achieved the best performance.

In order for the experiments to be deterministic in nature, we have configured Hugin to use random seeds for all the model and dataset initialization methods.

Training was performed on the proposed dataset that was split by using the Stratified K-folds algorithm (Section~\ref{sec:kfolds}). The folds were created previous to the training process and stored in a separate group in the zarr format (Section~\ref{sec:my-dataset}).

The different model configurations were trained using data ranging from the weeks 14-40 of the year 2018 (from April to September) with the reasoning being that of having the most samples containing vegetation in different states.

Samples of YAML configuration files for Hugin can be found within Appendix~\ref{sec:hugin-configs}.


%% file: sections/experiments.tex
\chapter{Experiments}\label{ch:experiments}

The current chapter presents results following training multiple~\gls{cnn} configurations (of topologies discussed in Sections~\ref{sec:segnet},~\ref{sec:unet},~\ref{sec:resnet}) using the dataset proposed in Section~\ref{sec:my-dataset} split using the K-folds~\ref{sec:kfolds} algorithm. The models were trained for semantic segmentation of vegetation using classes included in the~\gls{clc} inventory.

Table~\ref{tab:model-performance-metrics} shows the performance of the trained models for the testing dataset. While adding a significant time overhead in order to converge, we can notice that SegNet does not yield satisfactory results being in fact, the worst overall model.

Altough there exist a high number of training samples in the weeks that were selected for training (April to September) which include a small overall cloud cover percentage, and U-Net is a model that is designed to perform on small datasets, it has given the best performance overall, taking into account the~\gls{miou} metric.

\begin{table}[h!]
  \begin{center}
    \begin{tabular}{lcccccr}
      \textbf{Model} & \textbf{accuracy} & \textbf{precision} & \textbf{recall} & \textbf{MIoU} \\
      \hline
      U-Net  & 0.772349 & 0.918345 & 0.500243 & 0.602314 \\
      U-Net  & 0.682321 & 0.921134 & 0.499213 & 0.553526 \\
      SegNet & 0.425534 & 0.218452 & 0.323334 & 0.243481 \\
      ResNet & 0.525591 & 0.748259 & 0.402512 & 0.288468 \\
      ResNet & 0.588212 & 0.748592 & 0.448209 & 0.523451 \\
    \end{tabular}
  \end{center}
  \caption{Performance metrics of trained models.}
  \label{tab:model-performance-metrics}
\end{table}

The training time for both U-Net and ResNet was almost half of the convergence time of SegNet, while yielding better performance using the same configuration (the Adam optimizer and Categorical cross-entropy as a loss function) with the same initialization parameters and the same inference of data.

\begin{figure}[h!]
  \begin{center}
    \includegraphics[scale=0.3]{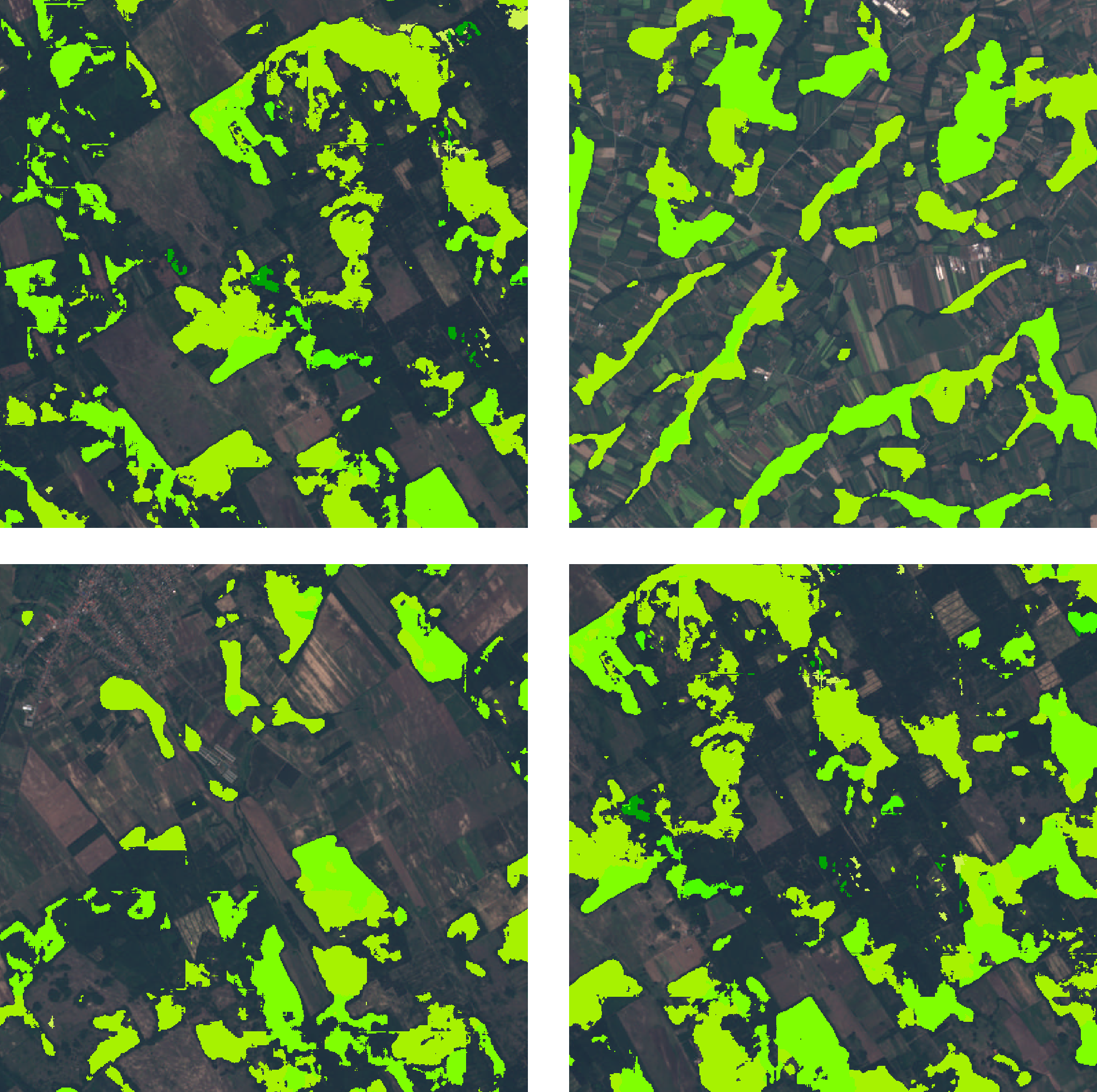}
    \caption{Output segmentation masks from the trained models\\U-Net(left) and ResNet (right)}
    \label{fig:output-segmentations}
  \end{center}
\end{figure}

Illustrated in Figure~\ref{fig:output-segmentations} are segmentation masks from U-Net and ResNet (both models for each network). Visually, we can notice that both trained topologies are able to produce segmentation masks, altough with problems: some regions include sparse segmentation masks even if vegetation is still visible. For further correction of these masks postprocessing techniques in order to fill the sparse masks can be employed in order for the segmentation to be more complete. This can be performed using methods that include nearest-neighbor sampling or expansion of the masks.

%% file: sections/conclusions.tex
\chapter{Conclusions}

\paragraph*{Summary}

\mbox{} \\

The current work has shown inner workings of Artificial Intelligence techniques based on Convolutional Neural Networks. Specifics of Remote Sensing and ways of processing data were also presented. We have highlighted a couple of existing datasets containing Remote Sensing data such as BigEarthNet (Section~\ref{sec:bigearthnet}), SpaceNet (Section~\ref{sec:spacenet}), Corine Land Cover (Section~\ref{sec:clc}), \gls{modis} Land Cover (Section~\ref{sec:modislc}) and LUCAS (Section~\ref{sec:lucas}). Specifics on building and training Machine Learning models such as common layers found in Convolutional Neural Networks (Section~\ref{sec:ml-op}), activation functions for activation of perceptrons (Section~\ref{sec:activation-funcs}), optimization methods (Section~\ref{sec:optimizers}) and also topologies of commonly used Fully Convolutional Networks (Sections~\ref{sec:segnet},~\ref{sec:unet},~\ref{sec:resnet}). Metrics for analyzing the performance of~\gls{ml} models are presented in~\ref{sec:performance}. Problems that often arise when dealing with Computer Vision tasks were discussed in Section~\ref{sec:cv} but also ways of mitigating such problems. Techniques for performing Multi-modal learning are presented within Section~\ref{sec:multi-modal}.

A proposal for creating a new, multi-modal and spatio-temporal dataset that is relevant in the current context, providing data that is useful for solving the task of semantic segmentation for land cover/land use. With the aggregation of Sentinel-2~\gls{msi} and Sentinel-3~\gls{olci} imagery, using Corine Land Cover~\gls{clc} for ground truth information. The value added by this dataset is validated by showing how the task of semantic segmentation of vegetation can be solved with Convolutional Neural Networks by learning vegetation features from the proposed dataset.


\paragraph*{Contributions}

\mbox{} \\

Given the current context of research in the fields of Remote Sensing and Machine Learning, the present work provides two important contributions: First, the creation of a new multi-modal and spatio-temporal dataset containing land cover information, paired with Sentinel-2~\gls{msi} and Sentinel-3~\gls{olci} data. Second, the validation of the dataset by providing information regarding performance of models trained to solve semantic segmentation of vegetation by using the proposed dataset.

\paragraph*{Open problems}

\mbox{} \\

Altough the results of experimenting with models trained for semantic segmentation of vegetation shown in Chapter~\ref{ch:experiments} provide some important results taking into account visual confirmation as proof of work as well, there are still open problems that interfere with the development of automated techinques based on Machine Learning.

Occlusion is one of the main problems that arise when dealing with Remote Sensing imagery, particularly when RGB data is used. Even if cloud formations can be identified by using processors such as Sen2cor (Section \ref{sec:cloudmasks}), they appear very frequently in Remote Sensing data. This unfortunately means that the number of samples we can reliably take into account when training Machine Learning models on such data is significantly reduced. More so, by observation we noticed that cloud formations appear more often during spring and autumn, therefore interferring with samples from these seasons, having to remove them from the training of models.

\paragraph*{Future work}

\mbox{} \\

Further ehnahcements for the proposed dataset include the addition of Sentinel-1~\gls{sar} data for providing another modality of learning. Another benefit could come from adding other land cover inventories such as~\gls{modis} land cover and~\gls{lucas}, but this also increases the complexity of translation and alignment of data while also adding a significant processing overhead.~\gls{clc} was chosen due to the granularity and multitude of land coverage classes it offers. Correlating Sentinel data with other inventories in the dataset could be proven as a benefit, but this implies in some cases the reduction of land cover classes since~\gls{modis} and~\gls{lucas} provide less in terms of classification schemes.

Additionally, including other cloud masks (Section~\ref{sec:cloudmasks}) generated by the MAJA\footnote{\url{https://labo.obs-mip.fr/multitemp/maccs-how-it-works/}} and FMask\footnote{\url{https://github.com/GERSL/Fmask}} algorithms can further help with selecting relevant samples from the dataset which are not occluded by cloud formations.



%% file: sections/appendix.tex
\appendix

\chapter{Remote Sensing}

\section*{Passive and active Remote Sensing}

\begin{figure}[h!]
  \begin{center}
    \includegraphics[scale=0.25]{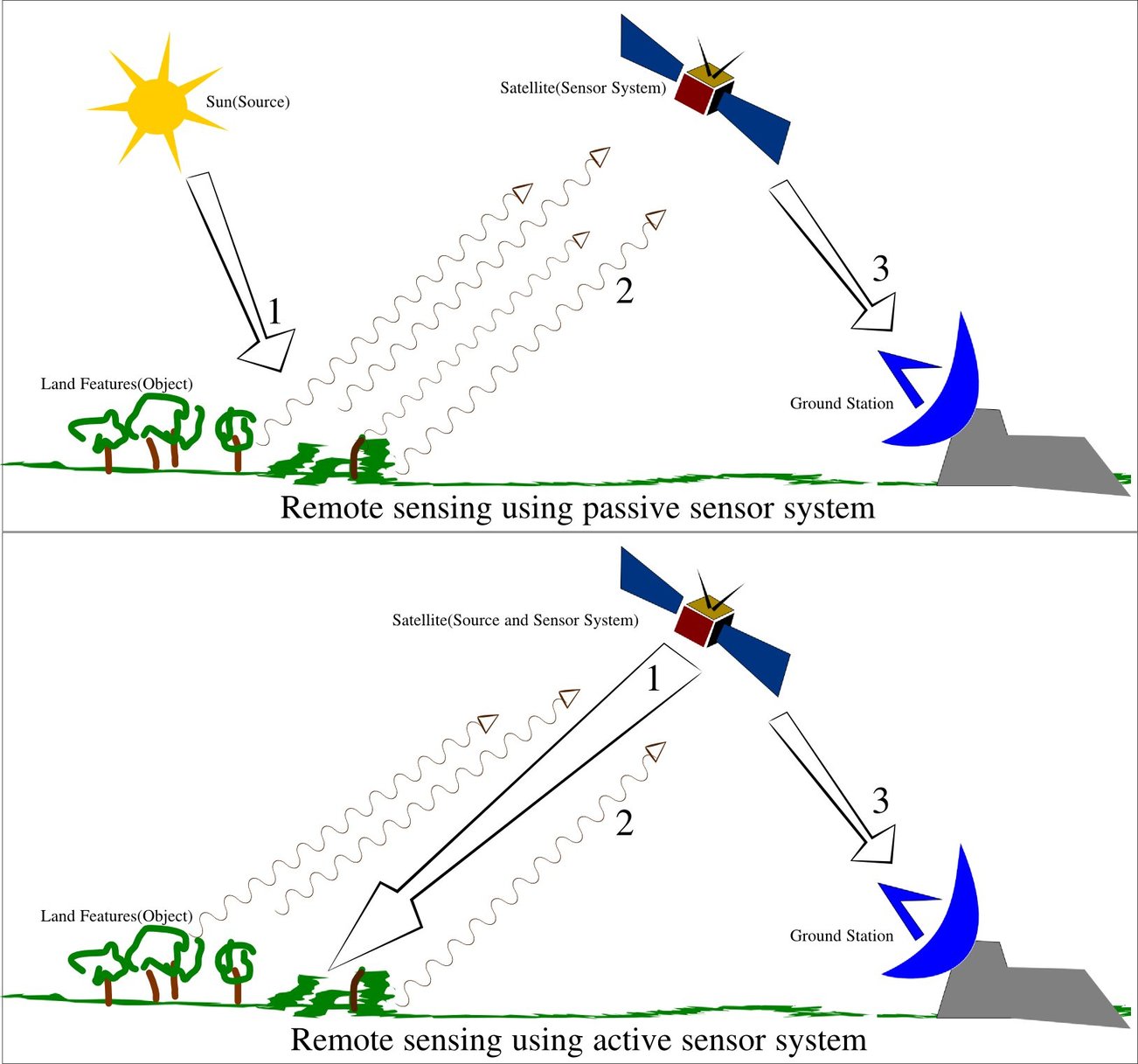}
    \caption{Difference the between active and passive Remote Sensing\\\scriptsize{Arkarjun, CC BY-SA 3.0 <https://creativecommons.org/licenses/by-sa/3.0>, via Wikimedia Commons}\\\scriptsize{\url{https://commons.wikimedia.org/wiki/File:Remote_Sensing_Illustration.jpg}}}
    \label{fig:active-passive-rs}
  \end{center}
\end{figure}

\pagebreak



\section*{Data regarding land coverage of Romania}\label{sec:landcover-info}



\begin{figure}[h!]
  \begin{center}
    \includegraphics[scale=2.3]{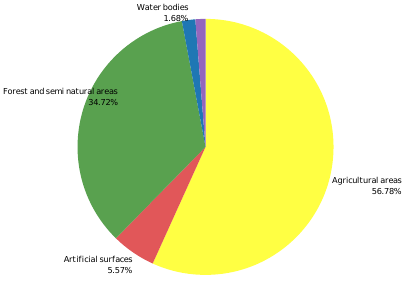}
    \captionsetup{justification=centering}
    \caption{Land cover in Romania as of 2018\\\tiny{Plot created with \url{https://land.copernicus.eu/dashboards/clc-clcc-2000-2018}}}
    \label{fig:land-cover-romania-2018}
  \end{center}
\end{figure}

Figure~\ref{fig:land-cover-romania-2018} showcases the breakdown of land coverage into the Corine Land Cover (\gls{clc}) Level-1 classes for the territory of Romania for the year 2018. We can notice that vegetation occupies 91.5\% (agricultural areas \& forest and semi-natural areas) of the land cover from the \gls{aoi} of the current work.

\section*{Using the~\gls{dhus} OpenSearch API}~\label{sec:dhus-opensearch}

The OpenSearch API provided by~\gls{dhus} instances can be queried for retrieving data using a cURL GET request:

\small
\begin{verbatim}
$curl --request GET 'https://scihub.copernicus.eu/dhus/api/stub/products
?filter=(
beginPosition:[2018-06-01T00:00:00.000Z TO 2018-09-01T23:59:59.999Z] AND
endPosition:[2018-06-01T00:00:00.000Z TO 2018-09-01T23:59:59.999Z]) AND
((platformname:Sentinel-3 AND filename:S3B_* AND
 producttype:OL_1_EFR___ AND instrumentshortname:OLCI))
AND footprint: Intersects
               (POLYGON((16.58910503349143 43.400842665330345,
                         26.95841113834191 43.400842665330345,
                         26.95841113834191 49.09541206485471,
                         16.58910503349143 49.09541206485471,
                         16.58910503349143 43.400842665330345)))
&offset=0&limit=25&sortedby=ingestiondate&order=desc'
\end{verbatim}
\normalsize

\pagebreak

\section*{Contents of an Sentinel-3~\gls{olci} archive}~\label{sec:s3-data-olci}

\begin{verbatim}
geo_coordinates.nc
instrument_data.nc
Oa01_radiance.nc
Oa02_radiance.nc
Oa03_radiance.nc
Oa04_radiance.nc
Oa05_radiance.nc
Oa06_radiance.nc
Oa07_radiance.nc
Oa08_radiance.nc
Oa09_radiance.nc
Oa10_radiance.nc
Oa11_radiance.nc
Oa12_radiance.nc
Oa13_radiance.nc
Oa14_radiance.nc
Oa15_radiance.nc
Oa16_radiance.nc
Oa17_radiance.nc
Oa18_radiance.nc
Oa19_radiance.nc
Oa20_radiance.nc
Oa21_radiance.nc
qualityFlags.nc
removed_pixels.nc
S3A_OL_1_EFR____20181231T081701_20181231...LN1_O_NT_002-ql.jpg
tie_geo_coordinates.nc
tie_geometries.nc
tie_meteo.nc
time_coordinates.nc
xfdumanifest.xml
\end{verbatim}

\pagebreak

\section*{Contents of an Sentinel-2~\gls{msi} archive}~\label{sec:s2-data-msi}

\footnotesize
\dirtree{%
.1 AUX\_DATA.
.2 DATASTRIP.
.3 DS\_CREO\_20210115T195324\_S20180831T133222.
.4 MTD\_DS.xml.
.4 QI\_DATA.
.1 GRANULE.
.2 L2A\_T22LGL\_A016666\_20180831T133222.
.3 AUX\_DATA.
.4 AUX\_ECMWFT.
.3 IMG\_DATA.
.5 R10m.
.6 T22LGL\_20180831T133221\_AOT\_10m.jp2.
.6 T22LGL\_20180831T133221\_B02\_10m.jp2.
.6 \ldots.
.6 T22LGL\_20180831T133221\_TCI\_10m.jp2.
.6 T22LGL\_20180831T133221\_WVP\_10m.jp2.
.5 R20m.
.6 T22LGL\_20180831T133221\_AOT\_20m.jp2.
.6 T22LGL\_20180831T133221\_B02\_20m.jp2.
.6 \ldots.
.6 T22LGL\_20180831T133221\_B8A\_20m.jp2.
.6 T22LGL\_20180831T133221\_SCL\_20m.jp2.
.6 T22LGL\_20180831T133221\_TCI\_20m.jp2.
.6 T22LGL\_20180831T133221\_WVP\_20m.jp2.
.5 R60m.
.6 T22LGL\_20180831T133221\_AOT\_60m.jp2.
.6 \ldots.
.6 T22LGL\_20180831T133221\_B8A\_60m.jp2.
.6 T22LGL\_20180831T133221\_SCL\_60m.jp2.
.6 T22LGL\_20180831T133221\_TCI\_60m.jp2.
.6 T22LGL\_20180831T133221\_WVP\_60m.jp2.
.3 MTD\_TL.xml.
.4 QI\_DATA.
.5 FORMAT\_CORRECTNESS.xml.
.5 GENERAL\_QUALITY.xml.
.5 GEOMETRIC\_QUALITY.xml.
.5 MSK\_CLDPRB\_20m.jp2.
.5 MSK\_CLDPRB\_60m.jp2.
.5 MSK\_CLOUDS\_B00.gml.
.5 MSK\_DEFECT\_B01.gml.
.5 MSK\_DEFECT\_B02.gml.
.5 MSK\_DEFECT\_B03.gml.
.5 \ldots.
.5 MSK\_DETFOO\_B01.gml.
.5 MSK\_DETFOO\_B02.gml.
.5 MSK\_DETFOO\_B03.gml.
.5 \ldots.
.5 MSK\_NODATA\_B01.gml.
.5 MSK\_NODATA\_B02.gml.
.5 MSK\_NODATA\_B03.gml.
.5 \ldots.
.5 MSK\_SATURA\_B01.gml.
.5 MSK\_SATURA\_B02.gml.
.5 MSK\_SATURA\_B03.gml.
.5 \ldots.
.5 MSK\_SNWPRB\_20m.jp2.
.5 MSK\_SNWPRB\_60m.jp2.
.5 MSK\_TECQUA\_B01.gml.
.5 MSK\_TECQUA\_B02.gml.
.5 MSK\_TECQUA\_B03.gml.
.5 \ldots.
.5 SENSOR\_QUALITY.xml.
.5 T22LGL\_20180831T133221\_PVI.jp2.
.1 INSPIRE.xml.
.1 manifest.safe.
.1 MTD\_MSIL2A.xml.
.1 rep\_info.
.2 S2\_PDI\_Level-2A\_Datastrip\_Metadata.xsd.
.2 S2\_PDI\_Level-2A\_Tile\_Metadata.xsd.
.2 S2\_User\_Product\_Level-2A\_Metadata.xsd.
.1 S2A\_MSIL2A\_20180831T133221\_N0213\_R081\_T22LGL\_20210115T195322-ql.jpg.
}
\normalsize

\pagebreak

\chapter{Sentinel-3 GPT preprocessing}\label{sec:gpt-processing}

\vspace{-1cm}

Using the Graph Processing Tool (\gls{gpt}) from ESA's \gls{snap}, we define an XML configuration of a pipeline that reprojects the data in an \href{https://epsg.io/4326}{EPSG:4326} projection, extracting the bands we desire, placing them into a single GeoTIFF file:

\begin{minted}[linenos,
               fontsize=\footnotesize,
               numbersep=5pt,
               frame=lines]{xml}
<graph id="Graph">
  <version>1.0</version>
  <node id="Reproject">
    <operator>Reproject</operator>
    <sources>
      <sourceProduct>${sourceProduct}</sourceProduct>
    </sources>
    <parameters>
      <crs>EPSG:4326</crs>
      <resampling>Nearest</resampling>
      <noDataValue>NaN</noDataValue>
      <includeTiePointGrids>true</includeTiePointGrids>
    </parameters>
  </node>
    <node id="BandsExtract">
    <operator>BandsExtractorOp</operator>
    <sources>
      <sourceProduct refid="Reproject"/>
    </sources>
    <parameters>
    <sourceBandNames>
      Oa01_radiance,Oa02_radiance, ..., Oa21_radiance
    </sourceBandNames>
      <sourceMaskNames></sourceMaskNames>
    </parameters>
  </node>
  <node id="Write_1">
    <operator>Write</operator>
    <sources>
      <sourceProduct refid="BandsExtract"/>
    </sources>
    <parameters>
      <file>${targetFile}</file>
      <formatName>GeoTIFF-BigTIFF</formatName>
    </parameters>
  </node>
</graph>
\end{minted}

\chapter{Hugin configuration examples}~\label{sec:hugin-configs}

\vspace{-1.3cm}

\begin{minted}[linenos,
               fontsize=\footnotesize,
               numbersep=5pt,
               frame=lines]{yaml}
configuration:
  model_path: "s3://sagegroup-data/alex/models/{name}"
  name: "unet6-clc-vegetation"
  workspace: "/home/alex/models/{name}"

data_source: !!python/object/apply:hugin.io.ZarrArrayLoader
  kwds:
    source: "s3://sagegroup-data/zarr/romania/2018/"
    flat: False
    slice_timestamps: [14, 32]
    inputs:
      input_1:
        component: "Sentinel-2/10m"
      input_2:
        component: "Sentinel-2/presence_mask"
    targets:
      output_1:
        component: "labels/clc/clc_10m"
    split_train_index_array: "labels/clc/samples/clc_10m/classes"
    split_test_index_array: "labels/clc/samples/clc_10m/classes"
    randomise: True
    random_seed: 42

trainer: !!python/object/apply:hugin.engine.scene.ArrayModelTrainer
  kwds:
    name: unet_clc_vegetation_2018
    model: !!python/object/apply:hugin.engine.keras.KerasModel
      kwds:
        name: keras_model
        model_builder: hugin.models.unet.unet3:unet
        swap_axes: True
        random_seed: 42
        model_builder_options:
          output_channels: 8
        model_path: "/data/bid/storage0/sage-storage/homes/sage/alex/models/{name}"
        batch_size: 1
        epochs: 100
        metrics:
          - !!python/object/apply:hugin.tools.utils.MultilabelMeanIOU
            kwds: { name: 'MeanIOU', num_classes: 8 }
        loss: categorical_crossentropy
        checkpoint:
          monitor: val_loss
        enable_multi_gpu: False
        num_gpus: 1
        use_multiprocessing: False
        workers: 12
        max_queue_size: 6
        optimizer: !!python/object/apply:tensorflow.keras.optimizers.Adam
          kwds:
            lr: !!float 0.001 # 0.0001
            beta_1: !!float 0.9
            beta_2: !!float 0.999
            epsilon: !!float 1e-7
        callbacks:
          - !!python/object/apply:tensorflow.keras.callbacks.EarlyStopping
            kwds:
              monitor: 'val_loss'
              min_delta: 0.001
              patience: 20
              verbose: 1
              mode: 'auto'
              baseline: null
              restore_best_weights: False
          - !!python/object/apply:tensorflow.keras.callbacks.ReduceLROnPlateau
            kwds:
              monitor: 'val_loss'
              patience: 5
              factor: !!float 0.2
\end{minted}

\chapter{Additional experiments results}

\begin{figure}[h!]
  \begin{center}
    \includegraphics[scale=0.38]{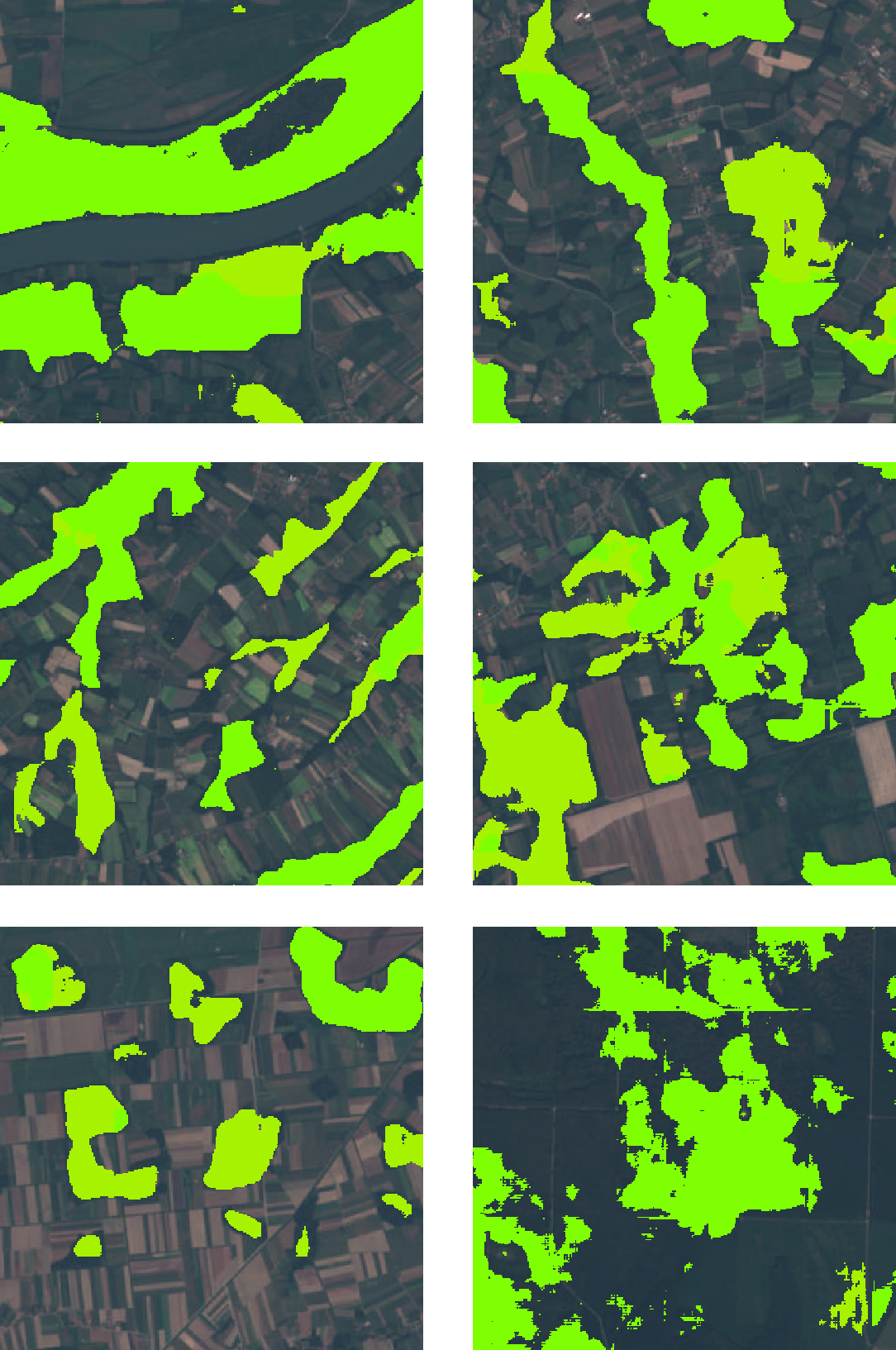}
    \captionsetup{justification=centering}
    \caption{Segmentation maps from the U-Net and ResNet models.}
    \label{fig:segmentation-maps-appendix}
  \end{center}
\end{figure}
